\documentclass{article}

\PassOptionsToPackage{numbers, compress, sort}{natbib}

\usepackage[final]{neurips_2024}
\usepackage{wrapfig} 
\usepackage{float} 
\usepackage[dvipsnames]{xcolor}
\usepackage{subcaption} 
\usepackage{varwidth}

\usepackage[utf8]{inputenc} 
\usepackage[T1]{fontenc}    
\usepackage{url}            
\usepackage[colorlinks=True, linkcolor=lb, citecolor=teal]{hyperref}
\usepackage{booktabs}       
\usepackage{amsfonts}       
\usepackage{nicefrac}       
\usepackage{microtype}      
\usepackage{xcolor}         

 \usepackage{enumitem}

\usepackage{graphicx}
\usepackage{subcaption} 

\usepackage{url}
\usepackage{enumitem} 
\usepackage{tabularx} 
\usepackage{centernot}
\usepackage[parfill]{parskip}
\usepackage{quiver}
\usepackage{bbm} 
\usepackage{graphicx} 
\usepackage{scalerel}
\usepackage{nicefrac} 
\usepackage{bbm} 

\usepackage{amsmath}
\usepackage{amssymb}
\usepackage{mathtools}
\usepackage{amsthm}

\DeclareMathOperator*{\argmin}{arg\,min} 

\usepackage[capitalize,noabbrev]{cleveref}

\theoremstyle{plain}

\newtheorem{proposition}{Proposition}
\newtheorem{lemma}{Lemma}
\theoremstyle{definition}
\newtheorem{definition}{Definition}

\theoremstyle{remark}

\theoremstyle{remark}

\newcommand{\RNum}[1]{\uppercase\expandafter{\romannumeral #1\relax}}

\definecolor{lb}{RGB}{31,119,180}
\usepackage{tcolorbox}
\usepackage{tabularx}
\usepackage{colortbl}
\tcbuselibrary{skins}

\tcbset{tab2/.style={enhanced,fonttitle=\bfseries,
colback=white,colframe=gray,colbacktitle=white,
coltitle=black,center title}}

\title{On Divergence Measures for Training  GFlowNets}

\newcommand{\pp}[1]{\vspace{2pt}\noindent\textbf{#1}}

\author{%
  Tiago da Silva \quad
  Eliezer de Souza da Silva \quad
  Diego Mesquita\\
\texttt{$\{$tiago.henrique, eliezer.silva, diego.mesquita$\}$@fgv.br}\\ 
  School of Applied Mathematics\\
  Getulio Vargas Foundation\\
  Rio de Janeiro, Brazil\\
}

\begin{document}

\maketitle

\begin{abstract}
  Generative Flow Networks (GFlowNets) are amortized inference models designed to sample from unnormalized distributions over composable objects, with applications in generative modeling for tasks in fields such as causal discovery, NLP, and drug discovery. Traditionally, the training procedure for GFlowNets seeks to minimize the expected log-squared difference between a proposal (forward policy) and a target (backward policy) distribution, which enforces certain flow-matching conditions. While this training procedure is closely related to variational inference (VI), directly attempting standard Kullback-Leibler (KL) divergence minimization can lead to proven biased and potentially high-variance estimators. Therefore, we first review four divergence measures, namely, Renyi-$\alpha$'s, Tsallis-$\alpha$'s, reverse and forward KL's, and design statistically efficient estimators for their stochastic gradients in the context of training GFlowNets. Then, we verify that properly minimizing these divergences yields a provably correct and empirically effective training scheme, often leading to significantly faster convergence than previously proposed optimization. To achieve this, we design control variates based on the REINFORCE leave-one-out and score-matching estimators to reduce the variance of the learning objectives' gradients. Our work contributes by narrowing the gap between GFlowNets training and generalized variational approximations, paving the way for algorithmic ideas informed by the divergence minimization viewpoint. \end{abstract}

\section{Introduction} 

The approximation of intractable distributions is one of the central issues in machine learning and modern statistics \cite{Jrvenp2023, Blei2017}. In reinforcement learning (RL), a recurring goal is to find a diverse set of high-valued state–action trajectories according to a reward function. This problem may be cast as sampling trajectories proportionally to the reward, which is generally an intractable distribution over the environment \cite{DBLP:levine-rl-control-inf, Bengio2021, buesing2020approximate, DBLP:journals/ml/KappenGO12-optimcinf}. Similarly, practical Bayesian inference and probabilistic models computations involve assessing intractable posterior distributions \cite{DBLP:journals/ml/JordanGJS99,yedidia2001bethe, robert2007bayesian}. 
In the variational inference (VI) approach, circumventing this intractability involves searching for a tractable approximation to the target distribution within a family of parametric models. Conventionally, the problem reduces to minimizing a divergence measure, such as Kullback-Leibler (KL) divergence \cite{DBLP:journals/ml/JordanGJS99, DBLP:journals/ftml/WainwrightJ08, Blei2017} or Renyi-$\alpha$ divergence \cite{li2016renyi, poczos11a}, between the variational approximation and the target. 

In particular, Generative Flow Networks (GFlowNets) \cite{Bengio2021, Foundations, theory} are a recently proposed family of variational approximations well-suited for distribution over compositional objects (e.g., graphs and texts). GFlowNets have found empirical success within various applications from causal discovery \cite{deleu2022bayesian, deleu2023joint}, NLP \cite{hu2023amortizing}, and chemical and biological modeling \cite{sequence, Bengio2021}. In a nutshell, a GFlowNet learns an iterative generative process (IGP) \cite{garipov2023compositional} over an extension of the target's support, which, for sufficiently expressive parameterizations of transition kernels, yields independent and correctly distributed samples \cite{Bengio2021, theory}. Remarkably, training GFlowNets typically consists of minimizing the log-squared difference between a proposal and target distributions over the extended space via SGD \cite{malkin2022trajectory, Foundations}, contrasting with divergence-minimizing algorithms commonly used in VI \cite{DBLP:conf/nips/RanganathTAB16, Blei2017}. 

Indeed, \citet{malkin2023gflownets} suggests that trajectory balance (TB) loss training for GFlowNets leads to better approximations of the target distribution than directly minimizing the reverse and forward KL divergence, particularly in setups with sparser rewards. Nevertheless, as we highlight in \autoref{sec:d}, these results are a potential consequence of biases and high variance in gradient estimates for the divergence's estimates, which can be overlooked in the evaluation protocol reliant upon sparse target distributions. Therefore, in \autoref{sec:e}, we present a comprehensive empirical investigation of the minimization of well-known \textit{f}-divergence measures (including reverse and forward KL), showing it is an effective procedure that often accelerates the training convergence of GFlowNets relative to alternatives. To achieve these results, we develop in \autoref{sec:c} a collection of control variates (CVs) \cite{mcbook, ranganath14} to reduce the variance without introducing bias on the estimated gradients, improving the efficiency of the optimization algorithms \cite{richter2020vargrad, shi2022gradient}.
In summary, our \emph{main contributions} are: 

\begin{enumerate}[itemsep=-2pt, leftmargin=.4cm] 
    \item We evaluate the performance of forward and reverse KL- \cite{kullback1951}, Renyi-$\alpha$ \cite{renyi1961measures} and Tsallis-$\alpha$ \cite{tsallis1988possible} divergences as learning objectives for GFlowNets through an extensive empirical campaign and highlight that they frequently outperform traditionally employed loss functions. 
    \item We design control variates for the gradients of GFlowNets' divergence-based objectives. Therefore, it is possible to perform efficient evaluations  
 of the optimization objectives using automatic differentiation frameworks \cite{paszke2019pytorch}, and the resulting experiments showcase the significant reduction in the variance of the corresponding estimators. 
    \item We developed a theoretical connection between GFlowNets and VI beyond the setup of finitely supported measures \cite{DBLP:journals/tmlr/ZimmermannLMN23, malkin2023gflownets}, establishing results for arbitrary topological spaces.  
\end{enumerate}

\section{Revisiting the relationship between GFlowNets and VI} \label{sec:p} 

Initially, we review \citet{theory}'s work on GFlowNets for distributions on topological spaces, a perspective applied consequentially to obtain the equivalence between GFlowNets training and VI divergence minimization in a more generic setting. Finally, we describe standard variance reduction techniques for solving stochastic optimization problems.  

\noindent\textbf{Notations.} Let $(\mathcal{S}, \mathcal{T})$ be a topological space with topology $\mathcal{T}$ and $\Sigma$ be the corresponding Borel $\sigma$-algebra. Also, let $\nu \colon \Sigma \rightarrow \mathbb{R}_{+}$ be a measure over $\Sigma$ and $\kappa_{f}, \kappa_{b} \colon \mathcal{S} \times \Sigma \rightarrow \mathbb{R}_{+}$ be transition kernels over $\mathcal{S}$. For each $(B_{1}, B_{2}) \in \Sigma \times \Sigma$, we denote by $\nu \otimes \kappa(B_{1}, B_{2}) \coloneqq \int_{B_{1}} \nu(\mathrm{d}s) k(s, B_{2})$. Likewise, we recursively define the \emph{product kernel} as $\kappa^{\otimes 0}(s, \cdot) = \kappa(s, \cdot)$ and, for $n \ge 1$, $\kappa^{\otimes n}(s, \cdot) = \kappa^{\otimes n - 1}(s, \cdot) \otimes \kappa$ for a transition kernel $\kappa$ and $s \in \mathcal{S}$. Note, in particular, that $\kappa^{\otimes n}$ is a function from $\mathcal{S} \times \Sigma^{\otimes n + 1}$ to $\mathbb{R}_{+}$, with $\Sigma^{\otimes n + 1}$ representing the product $\sigma$-algebra of $\Sigma$ \cite{david, Axler2020}. Moreover, if $\mu$ is an absolutely continuous measure relatively to $\nu$, denoted $\mu \ll \nu$, we write $\nicefrac{\mathrm{d}\mu}{\mathrm{d}\nu}$ for the corresponding density (Radom-Nikodym derivative) \cite{Axler2020}. Furthermore, we denote by $\mathcal{P}(A) = \{S \colon S \subseteq A\}$ the power-set of a set $A \subset \mathcal{S}$ and by $[d] = \{1, \dots, d\}$ the first $d$ positive integers. 

\noindent\textbf{GFlowNets.} A GFlowNet is, in its most general form, built upon the concept of a \emph{measurable pointed directed acyclic graph} (DAG) \cite{theory}, which we define next. Intuitively, it extends the notion of a \emph{flow network} to arbitrary measurable topological spaces, replacing the directed graph with a transition kernel specifying how the underlying states are connected. 

\begin{definition}[Measurable pointed DAG \cite{theory}] \label{def:aaa}
    Let $(\bar{\mathcal{S}}, \mathcal{T}, \Sigma)$ be a measurable topological space endowed with a reference measure $\nu$ and forward $\kappa_{f}$ and backward $\kappa_{b}$ kernels. Also, let $s_{o} \in \bar{\mathcal{S}}$ and $s_{f} \in \bar{\mathcal{S}}$ be distinguished elements in $\bar{\mathcal{S}}$, respectively called \emph{initial} and \emph{final} states, and $\mathcal{S} = \bar{\mathcal{S}} \setminus \{s_{f}\}$. A \emph{measurable pointed DAG} is then a tuple $(\mathcal{S}, \mathcal{T}, \Sigma, \kappa_{f}, \kappa_{b}, \nu)$ satisfying the properties below.
    \begin{enumerate}[itemsep=0pt,leftmargin=16pt]
        \item \textbf{(Terminality)} If $\kappa_{f}(s, s_{f}) >  0$, then $\kappa_{f}(s, s_{f}) = 1$ $\forall s \in \bar{\mathcal{S}}$. Also, $\kappa_{f}(s_{f}, \cdot) = \delta_{s_{f}}$. 
        \item \textbf{(Reachability)} For all $B \in \Sigma$, $\exists \, n \in \mathbb{N}$ s.t. $\kappa_{f}^{\otimes n}(s_{o}, B) > 0$, i.e., $B$ is reachable from $s_{o}$.
        \item \textbf{(Consistency)} For every $(B_{1}, B_{2}) \in \Sigma \times \Sigma$ such that $(B_{1}, B_{2}) \notin \{(s_{o}, s_{o}), (s_{f}, s_{f})\}$, $\nu \otimes \kappa_{f}(B_{1}, B_{2}) = \nu \otimes \kappa_{b}(B_{2}, B_{1})$. Moreover, $\kappa_{b}(s_{o}, B) = 0$ for every $B \in \Sigma$. 
        \item \textbf{(Continuity)} $s \mapsto \kappa_{f}(s, B)$ is continuous for $B \in \Sigma$.
        \item \textbf{(Finite absorption)} There is a $N \in \mathbb{N}$ such that $\kappa_{f}^{\otimes N}(s, \cdot) = \delta_{s_{f}}$ for every $s \in \mathcal{S}$. We designate the corresponding DAG as \emph{finitely absorbing}.  
    \end{enumerate}
\end{definition}

In this setting, the elements in the set $\mathcal{X} = \{s \in \mathcal{S} \setminus \{s_{f}\} \colon \kappa_{f}(s, \{s_{f}\}) > 0 )\}$ are called \emph{terminal states}. Illustratively, when $\mathcal{S}$ is finite and $\nu$ is the counting measure, the preceding definition corresponds to a connected DAG with an edge from $s \in \mathcal{S}$ to $s' \in \mathcal{S}$ iff $\kappa_{f}(s, \{s'\}) > 0$, with condition 5) ensuring acyclicity and condition 2) implying connectivity. A GFlowNet, then, is characterized by a measurable pointed DAG, a potentially unnormalized distribution over terminal states $\mathcal{X}$ and learnable transition kernels on $\mathcal{S}$ 
 (\autoref{def:aa}). Realistically, its goal is to find an IGP over $\mathcal{S}$ which, starting at $s_{o}$, samples from $\mathcal{X}$ proportionally to a given positive function. 

\begin{definition}[GFlowNets \cite{theory}] \label{def:aa} 
    A \emph{GFlowNet} is a tuple $(\mathcal{G}, \! P_{F}, \! P_{B}, \! \mu)$ composed of a measurable pointed DAG $\mathcal{G}$, a $\sigma$-finite measure $\mu \ll \nu$, and $\sigma$-finite Markov kernels $P_{F} \ll \kappa_{f}$ and $P_{B} \ll \kappa_{b}$, respectively called \emph{forward} and \emph{backward} policies.  
\end{definition}
  
\noindent\textbf{Training GFlowNets.} In practice, we denote by $p_{F_\theta} \colon \mathcal{S} \times \mathcal{S} \rightarrow \mathbb{R}_{+}$  the density of $P_{F}$ relative to $\kappa_{f}$, which we parameterize using a neural network with weigths $\theta$. Similarly, we denote by $p_B$ the density of $P_B$ wrt $k_b$. Our objective is, for a given \emph{target measure} $R \ll \mu$ on $\mathcal{X}$ with $r = \nicefrac{\mathrm{d}R}{\mathrm{d}\mu}$, estimate the $\theta$ for which the distribution over $\mathcal{X}$ induced by $P_{F}(s_{o}, \cdot)$ matches $R$, i.e., for every $B \in \Sigma$, 
\begin{equation*}
    {\textstyle \sum_{n \ge 0}}  \int_{\mathcal{S}^{n}} p_{F_\theta}^{\otimes n}(s_{o}, s_{1:n}, s_{f}) \mathbbm{1}_{s_{n} \in B} \kappa_{f}^{\otimes n}(s_{o}, \mathrm{d}s_{1:n}) = \frac{R(B)}{R(\mathcal{X})}. 
\end{equation*}
\noindent Importantly, the above sum contains only finitely many non-zero terms due to the finite absorption property of $\kappa_{f}$. To ensure that $p_{F_\theta}$ abides by this equation, Lahlou et al. \cite{theory} showed it suffices that one of the next \emph{balance conditions} are concomitantly satisfied by $P_{F}$ and $P_{B}$. 
\begin{definition}[Trajectory balance condition] For all $n \ge 0$ and $\mu^{\otimes n}$-almost surely $\forall s_{1:n} \in \mathcal{S}^{n}$, $p_{F_\theta}^{\otimes n}(s_{o}, s_{1:n}, s_{f}) \! = \! \frac{r(s_{n})}{Z_\theta} p_{B}^{\otimes n}(s_{n}, s_{n:1}, s_{o})$, w/ $Z_\theta$ denoting the target distribution's partition function. 
\end{definition}

\begin{definition}[Detailed balance condition]
        For an auxiliary parametric function $u \colon \mathcal{S} \rightarrow \mathbb{R}_{+}$ and $\mu^{\otimes 2}$-almost surely on $(s, s') \in \mathcal{S}^{2}$, $u(s) p_{F_\theta}(s, s')  = u(s') p_{B}(s', s)$ and $u(x) = r(x)$ for $x \in \mathcal{X}$. 
\end{definition}  
To enforce the trajectory balance (TB) or detailed balance (DB) conditions, we conventionally define a stochastic optimization problem to minimize the expected log-squared difference between the left- and right-hand sides of the corresponding condition under a probability measure $\xi$ supported on an appropriate space \cite{theory, malkin2022trajectory, Foundations, deleu2022bayesian, lau2023dgfn, stochastic, liu2023dropout}. For TB, e.g., we let $\xi$ be defined on $\Sigma^{\otimes N}$ with support $\text{supp}(\mu^{\otimes N})$. Then, we estimate the GFlowNet's parameters $\theta$ by minimizing 
\begin{equation} \label{eq:aa} 
    \underset{\tau \sim \xi}{\mathbb{E}} \left[  \left( \sum_{0 \le i \le N - 1} \left(\log \frac{p_{F_\theta}(s_{i}, s_{i + 1})}{p_{B}(s_{i + 1}, s_{i})}\right) + \log \frac{Z_\theta}{r(s_{h})} \right)^{2} \right] 
\end{equation}
with $\tau = s_{o:N}$ and $h = \max \{i \colon s_{i} \neq s_{f} \}$ being the last non-final state's index in $\tau$. As denoted by the subscript on $Z_\theta$, using the TB loss incurs learning the target's normalizing constant. While tuning $p_B$ during training is also possible, the common practice is to keep it fixed.

Henceforth, we will consider the measurable space of \emph{trajectories} $(\mathcal{P}_{\mathcal{S}}, \Sigma_{P})$, with $\mathcal{P}_{\mathcal{S}} = \{ (s, s_{1}, \dots, s_{n}, s_{f}) \in \mathcal{S}^{n + 1} \times \{s_{f}\} \colon 0 \le n \le N - 1 \}$ and $\Sigma_{P}$ as the $\sigma$-algebra generated by $\bigcup_{n=1}^{N + 1} \Sigma^{\otimes n}$. For notational convenience, we use the same letters for representing the measures and kernels of $(\mathcal{S}, \Sigma)$ and their natural product counterparts in $(\mathcal{P}_{\mathcal{S}}, \Sigma_{P})$, which exist by Carathéodory extension's theorem \cite{david}; for example, $\nu(B) = \nu^{\otimes n}(B)$ for $B = (B_{1}, \dots, B_{n}) \in \Sigma^{\otimes n}$ and $p_{F_\theta}(\tau | s_{o} ; \theta)$ is the density of $P_{F}^{\otimes n + 1}(s_{o}, \cdot)$ for $\tau = (s_{o}, s_{1}, \dots, s_{n}, s_{f})$ relatively to $\mu^{\otimes n}$. In this case, we will write $\tau$ for a generic element of $\mathcal{P}_{\mathcal{S}}$ and $x$ for its terminal state (which is unique by \autoref{def:aaa}).  
For a comprehensive overview of GFlowNets, please refer to \cite{malkin2022trajectory, theory}. 

\noindent\textbf{GFlowNets and VI.} GFlowNets can be interpreted as hierarchical variational models by framing the forward policy \( p_{F_\theta}(\tau | s_{o} ; \theta) \) in \( (\mathcal{P}_{\mathcal{S}}, \Sigma_{P}) \) as a proposal to \( \frac{r(x)}{Z} p_{B}(\tau | x) \). \citet{malkin2023gflownets} demonstrated that, for discrete target distributions, the TB loss in \eqref{eq:aa} aligns with the KL divergence in terms of expected gradients. Extending this, our \autoref{prop:aaa} establishes that this relationship also holds for distributions over arbitrary topological spaces.

\begin{proposition}[TB loss- and KL divergence gradients for topological spaces] \label{prop:aaa} 
    Let $\mathcal{L}_{TB}(\tau ; \theta) = \left(\log \nicefrac{Z p_{F_\theta}(\tau | s_{o}; \theta)}{r(x) p_{B}(\tau | x)}\right)^{2}$ and $p_{B}(\tau)  =  \frac{r(x)}{Z} p_{B}(s_{n-1:o} | x)$ for $\tau  = (s_{o}, \dots, s_{n - 1}, x, s_{f})$. Then, 
    \begin{equation} \label{eq:a}
        \nabla_{\theta} \mathbb{E}_{\tau \sim P_{F}(s_{o}, \cdot)} \left[ \mathcal{L}_{TB}(\tau ; \theta) \right] = 2 \nabla_{\theta} \mathcal{D}_{KL} [ P_{F} || P_{B} ], 
    \end{equation}
    where $\mathcal{D}_{KL} [ p_{F_\theta} || p_{B} ] = \mathbb{E}_{\tau \sim P_{F}(s_{o}, \cdot)} \left[ \log \nicefrac{p_{F_\theta}(\tau | s_{o}; \theta)}{p_{B}(\tau)}\right]$ is the KL divergence between $P_{F}$ and $P_{B}$. 
\end{proposition}



\noindent This proposition shows that minimizing the on-policy TB loss is theoretically comparable to minimizing the KL divergence between $P_{F}$ and $P_{B}$ in terms of convergence speed. Since the TB loss requires estimating the intractable $R(\mathcal{X})$, the KL divergence, which avoids this estimation, can be a more suitable objective. Our experiments in \autoref{sec:e} support this, with proofs provided in \autoref{app:proofs}. Extending this result to general topological spaces broadens the scope of divergence minimization strategies, extending guarantees for discrete spaces to continuous and mixed spaces. This generalization aligns with advances in generalized Bayesian inference \cite{DBLP:journals/jmlr/KnoblauchJD22} and generalized VI in function spaces \cite{DBLP:conf/nips/WildHS22}, via optimization of generic divergences. We make the method theoretically firm and potentially widely applicable by proving the equivalence in these broader contexts.

\noindent\textbf{Variance reduction.} A naive Monte Carlo estimator for the gradient in \autoref{eq:a} has high variance \cite{dieng2017variational}, impacting the efficiency of stochastic gradient descent \cite{Williams1992}. To mitigate this, we use \emph{control variates}—random variables with zero expectation added to reduce the estimator's variance without bias \cite{mcbook, ranganath14}. This method, detailed in \autoref{sec:c}, significantly reduces noise in gradient estimates and pragmatically improves training convergence, as shown in the experiments in \autoref{sec:e}.

\section{Divergence measures for learning GFlowNets} \label{sec:d}  

This Section presents four different divergence measures for training GFlowNets and the accompanying gradient estimators for stochastic optimization. Regardless of the learning objective, recall that our goal is to estimate $\theta$ by minimizing a discrepancy measure $D$ between $P_{F}$ and $P_{B}$ that is globally minimized if and only if $P_{F} = P_{B}$, i.e., 
\begin{equation} \label{eq:aaaaa} 
    \theta^{*} = \argmin_{\theta} D(P_{F}, P_{B}),  
\end{equation}
in which $P_{B}$ is typically fixed and $P_{F}$'s density $p_F^\theta$ is parameterized by $\theta$.    

\subsection{Renyi-$\alpha$ and Tsallis-$\alpha$ divergences} 

Renyi-$\alpha$ \cite{renyi1961measures} and Tsallis-$\alpha$ \cite{tsallis1988possible} are families of statistical divergences including, as limiting cases, the widespread KL divergence (\autoref{subsec:a}) \cite{minka2005divergence}; see \autoref{def:aaaa}. These divergences have been successfully applied to both variational inference \cite{li2016renyi} and policy search for model-based reinforcement learning \cite{depeweg2016learning}. Moreover, as we highlight in \autoref{sec:e}, their performance is competitive with, and sometimes better than, traditional learning objectives for GFlowNets based on minimizing log-squared differences between proposal and target distributions. 

\begin{definition}[Renyi-$\alpha$ and Tsallis-$\alpha$ divergences] \label{def:aaaa}
    Let $\alpha \in \mathbb{R}$. Also, let $p_{F_\theta}$ and  $p_{B}$ be GFlowNet's forward and backward policies, respectively. Then, the \emph{Renyi-$\alpha$ divergence} between $P_{F}$ and $P_{B}$ is 
    \begin{equation*}
        \mathcal{R}_{\alpha}(P_{F} || P_{B}) = \frac{1}{\alpha - 1} \log \int_{\mathcal{P}_{\mathcal{S}}} p_{F_\theta}(\tau | s_{o})^{\alpha} p_{B}(\tau)^{1 - \alpha} \kappa_{f}(s_{o}, \mathrm{d}\tau). 
    \end{equation*}
    Similarly, the \emph{Tsallis-$\alpha$ divergence} between $P_{F}$ and $P_{B}$ is 
    \begin{equation*}
        \mathcal{T}_{\alpha}(P_{F} || P_{B}) = \frac{1}{\alpha - 1} \left(\int_{\mathcal{P}_{\mathcal{S}}} p_{F_\theta}(\tau | s_{o})^{\alpha} p_{B}(\tau)^{1 - \alpha} \kappa_{f}(s_{o}, \mathrm{d}\tau) - 1 \right). 
    \end{equation*}
\end{definition}

\captionsetup[figure]{font=small}
\begin{wrapfigure}[14]{!ht}{.3\textwidth} 
    \centering
    \vspace{-21pt} 
    \includegraphics[width=\linewidth]{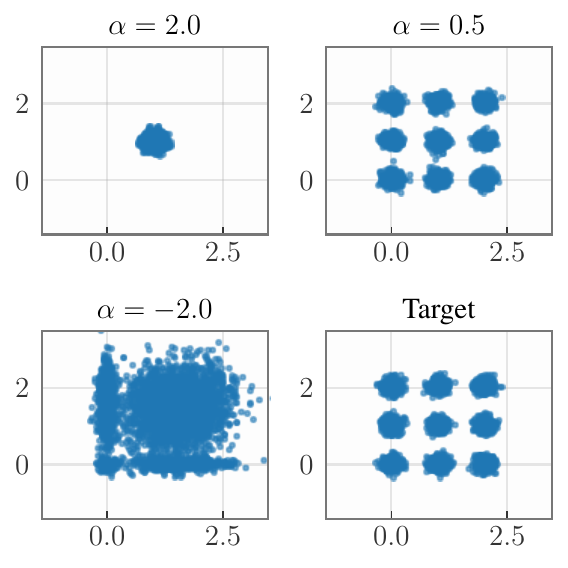}
    \caption{Mode-seeking ($\alpha = 2$) versus mass-covering ($\alpha = -2$) behaviour in $\alpha$-divergences.} 
    \label{fig:alpha}
\end{wrapfigure}
\captionsetup[figure]{font=normal}
From \autoref{def:aaaa}, we see that both Renyi-$\alpha$ and Tsallis-$\alpha$ divergences transition from a mass-covering to a mode-seeking behavior as $\alpha$ ranges from $-\infty$ to $\infty$. Regarding GFlowNet-training, this flexibility suggests that $\mathcal{R}_{\alpha}$ and $\mathcal{T}_{\alpha}$ are appropriate choices both, e.g., for carrying out Bayesian inference \cite{deleu2023joint} --- where interest lies in obtaining an accurate approximation to a posterior distribution ---, and for combinatorial optimization \cite{Zhang2023} --- where the goal is to find a few high-valued samples. Additionally, the choice of $\alpha$ provides a mechanism for controlling which trajectories are preferentially sampled during training, with larger values favoring the selection of trajectories leading to high-probability terminal states, resembling the effect of  $\varepsilon$-greedy \cite{malkin2022trajectory}, thompson-sampling \cite{rectorbrooks2023thompson}, local-search \cite{kim2023local}, and forward-looking \cite{pan2023generative, pan2023better} techniques for carrying out off-policy training of GFlowNets \cite{malkin2023gflownets}. 

To illustrate the effect of $\alpha$ on the learning dynamics of GFlowNets, we show in \autoref{fig:alpha} an \emph{early stage} of training to sample from a homogeneous mixture of Gaussian distributions by minimizing Renyi-$\alpha$ divergence for different values of $\alpha$; see \autoref{sec:env} for details on this experiment. At this stage, we note that the GFlowNet covers the target distribution's modes but fails to separate them when $\alpha$ is large and negative. In contrast, a large positive $\alpha$ causes the model to focus on a single high-probability region. Therefore, the use of an intermediate value for $\alpha = 0.5$ culminates in a model that accurately approximates the target distribution. Also, our early experiments suggested the persistence of such dependence on $\alpha$ for diverse learning tasks, with $\alpha = 0.5$ leading to the best results. Thus, we fix $\alpha = 0.5$ throughout our experimental campaign.  

Importantly, we need only the gradients of $\mathcal{R}_{\alpha}$ and $\mathcal{T}_{\alpha}$ for solving the optimization problem in \autoref{eq:aaaaa} and, in particular, learning the target distribution's normalizing constant is unnecessary, as we underline in the lemma below. This property distinguishes such divergence measures from both TB and DB losses in \autoref{eq:aa} and, in principle, simplifies the training of GFlowNets. 

\begin{lemma}[Gradients for $\mathcal{R}_{\alpha}$ and $\mathcal{T}_{\alpha}$] \label{lemma:a} 
    Let $\theta$ be the parameters of $p_{F_\theta}$ in \autoref{def:aaaa} and, for $\tau \in \mathcal{P}_{\mathcal{S}}$, $g(\tau, \theta) = \left( \nicefrac{p_{B}(\tau | x) r(x)}{p_{F_\theta}(\tau | s_{o} ; \theta)} \right)^{1 - \alpha}$. The gradient of $\mathcal{R}_{\alpha}$ wrt $\theta$ is 
    \begin{equation*}
        \nabla_{\theta} \mathcal{R}_{\alpha}(p_{F_\theta} || p_{B}) = 
        \frac
        {{\mathbb{E}}[ \nabla_{\theta} g(\tau, \theta) + g(\tau, \theta) \nabla_{\theta} \log p_{F_\theta}(\tau | s_{o} ; \theta)]}
        {(\alpha - 1) {\mathbb{E}} [ g(\tau, \theta) ] };  
    \end{equation*}
    the expectations are computed under $P_{F}$. Analogously, the gradient of $\mathcal{T}_{\alpha}$ wrt $\theta$ is 
    \begin{equation*}
        \nabla_{\theta} \mathcal{T}_{\alpha} (p_{F_\theta} || p_{B}) \stackrel{C}{=} \frac{\mathbb{E}[\nabla_{\theta} g(\tau, \theta) + g(\tau, \theta) \nabla_{\theta} \log p_{F_\theta}(\tau | s_{o}; \theta)]}{(\alpha - 1)}, 
    \end{equation*}
    in which $\stackrel{C}{=}$ denotes equality up to a multiplicative constant. 
\end{lemma}

\autoref{lemma:a} uses the REINFORCE method \cite{Williams1992} to compute the gradients of both $\mathcal{R}_{\alpha}$ and $\mathcal{T}_{\alpha}$, and we implement Monte Carlo estimators to approximate the ensuing expectations based on a batch of trajectories $\{\tau_{1}, \dots, \tau_{N}\}$ sampled during training \cite{malkin2023gflownets}. Also, note that the function $g$ is computed outside the log domain and, therefore, particular care is required to avoid issues such as numerical underflow of the unnormalized distribution \cite{tiapkin2024generative, Bengio2021}. 
In our implementation, we sample an initial batch of trajectories $\{\tau_{i}\}_{i=1}^{N}$ and compute the maximum of $r$ among the sampled terminal states in log space, i.e., $\log \tilde{r} = \max_{i} \log r(x_{i})$. Then, we consider $\log \tilde{r}(x) = \log r(x) - \log \tilde{r}$ as the target's unnormalized log density.  
In \autoref{sec:c}, we will consider the design of variance reduction techniques to decrease the noise level of gradient estimates and possibly speed up the learning process. 

\subsection{Kullback-Leibler divergence} \label{subsec:a}  

The KL divergence \cite{kullback1951} is a limiting member of the Renyi-$\alpha$ and Tsallis-$\alpha$ families of divergences, derived when $\alpha \rightarrow 1$ \cite{poczos11a}, and is the most widely deployed divergence measure in statistics and machine learning. To conduct variational inference, one regularly considers both the \emph{forward} and \emph{reverse} KL divergences, which we review in the definition below. 

\begin{definition}[Forward and reverse KL] 
    The \emph{forward} KL divergence between a target $P_{B}$ and a proposal $P_{F}$ is 
    $\mathcal{D}_{KL}[ P_{B} || P_{F} ] = \mathbb{E}_{\tau \sim P_{B}(s_{f}, \cdot)} \left[ \log \nicefrac{p_{B}(\tau)}{p_{F_\theta}(\tau | s_{o})} \right]$.   
    Also, the \emph{reverse} KL divergence is defined by 
    $\mathcal{D}_{KL}[ P_{F} || P_{B} ] = \mathbb{E}_{\tau \sim P_{F}(s_{o}, \cdot)} \left[ \log \nicefrac{p_{F_\theta}(\tau | s_{o})}{p_{B}(\tau)} \right]$. 
\end{definition}

Remarkably, we cannot use a simple Monte Carlo estimator to approximate the forward KL due to the presumed intractability of $P_{B}$ (which depends directly on $R$). As a first approximation, we could estimate $\mathcal{D}_{KL}[P_{B} || P_{F}]$ via importance sampling w/ $P_{F}$ as a proposal distribution as in \cite{malkin2023gflownets}: 
\begin{equation}
    \mathcal{D}_{KL}[P_{B} || P_{F}] = \mathbb{E}_{\tau \sim P_{F}} \left[ \frac{p_{B}(\tau)}{p_{F_\theta}(\tau | s_{o})} \log \frac{p_{B}(\tau)}{p_{F_\theta}(\tau | s_{o})} \right], 
\end{equation}
and subsequently implement a REINFORCE estimator to compute $\nabla_{\theta} \mathcal{D}_{KL}[P_{B} || P_{F}]$. Nevertheless, as we only need the divergence's derivatives to perform SGD, we 
apply the importance weights directly to the gradient estimator. We summarize this approach in the lemma below. 

\begin{lemma}[Gradients for the KL divergence] \label{lemma:aa} 
    Let $\theta$ be the parameters of $P_{F}$ and $s(\tau ; \theta) = \log p_{F_\theta}(\tau | s_{o} ; \theta)$. Then, the gradient of $\mathcal{D}_{KL}[P_{F} || P_{B}]$ relatively to $\theta$ satisfies  
    \begin{equation*}
            \nabla_{\theta} \mathcal{D}_{KL} \left[ P_{F} || P_{B} \right] = \mathbb{E}_{\tau \sim P_{F}(s_{o}, \cdot)} \left[ \nabla_{\theta} s (\tau ; \theta) + \log \frac{p_{F_\theta}(\tau | s_{o})}{p_{B}(\tau | x) r(x)} \nabla_{\theta} s(\tau ; \theta) \right]  
    \end{equation*}
    \noindent Correspondingly, the gradient of $\mathcal{D}_{KL}[P_{B} || P_{F}]$ wrt $\theta$ is 
    \begin{equation*}
        \nabla_{\theta} \mathcal{D}_{KL}[P_{B} || P_{F}] \stackrel{C}{=} - \mathbb{E}_{\tau \sim P_{F}(s_{o}, \cdot)} \left[ \frac{p_{F_\theta}(\tau | s_{o})}{p_{B}(\tau | x) r(x)} \nabla_{\theta} s(\tau ; \theta) \right]. 
    \end{equation*}
\end{lemma}

Crucially, choosing an appropriate learning objective is an empirical question that one should consider on a problem-by-problem basis --- similar to the problem of selecting among Markov chain simulation techniques \cite{geyer1991markov}. In particular, a one-size-fits-all solution does not exist; see \autoref{sec:e} for a thorough experimental investigation. Independently of the chosen method, however, the Monte Carlo estimators for the quantities outlined in \autoref{lemma:aa} are of potentially high variance and may require a relatively large number of trajectories to yield a reliable estimate of the gradients \cite{Williams1992}. The following sections demonstrate that variance reduction techniques alleviate this issue.

\section{Control variates for low-variance gradient estimation} \label{sec:c}   

\begin{figure*}
    \centering
    \includegraphics[width=.8\textwidth]{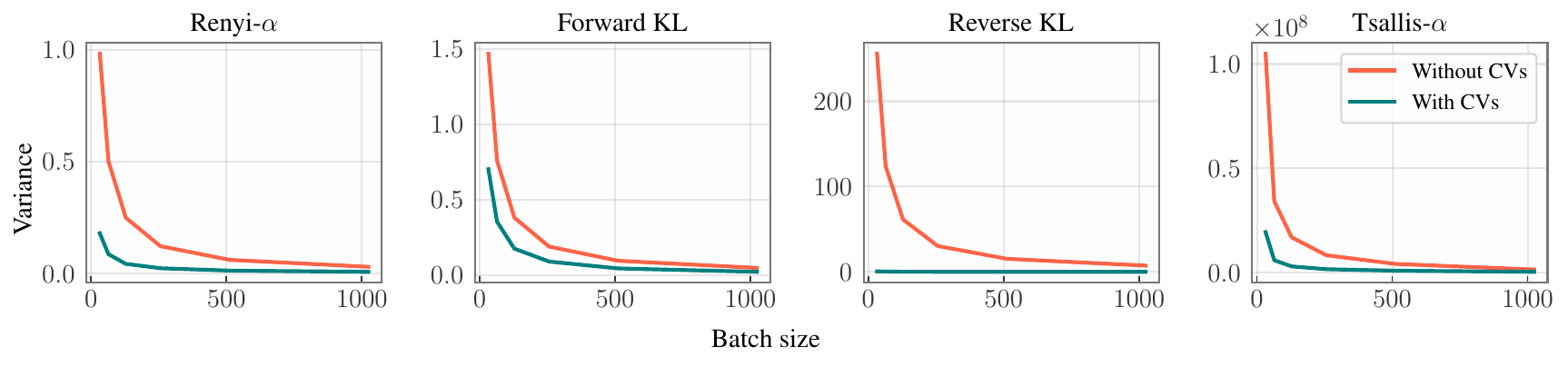}
    \caption{\textbf{Variance of the estimated gradients as a function of the trajectories' batch size.} Our control variates greatly reduce the estimator's variance, even for relatively small batch sizes.}
    \label{fig:cv}
\end{figure*}

\pp{Control variates.} We first review the concept of a control variate. Let $f \colon \mathcal{P}_{\mathcal{S}} \rightarrow \mathbb{R}$ be a real-valued measurable function and assume that our goal is to estimate $\mathbb{E}_{\tau \sim \pi} \left[ f(\tau) \right]$ according to a probability measure $\pi$ on $\Sigma_{P}$ (see \autoref{sec:p} to recall the definitions). The variance of a naive Monte Carlo estimator for this quantity is $\nicefrac{\text{Var}_{\pi}(f(\tau))}{n}$. On the other hand, consider a random variable (RV) $g \colon \mathcal{P}_{\mathcal{S}} \rightarrow \mathbb{R}$ positively correlated with $f$ and with known expectation $\mathbb{E}_{\pi}[g(\tau)]$. Then, the variance of a naive Monte Carlo for $\mathbb{E}_{\pi}\left[f(\tau) - a (g(\tau) - \mathbb{E}_{\pi}[g(\tau)])\right]$ for a \emph{baseline} $a \in \mathbb{R}$ is 
\begin{equation} \label{eq:aaaaaa} 
    \frac{1}{n} \left [ \text{Var}_{\pi} (f(\tau)) \! - \! 2 a \text{Cov}_{\pi} ( f(\tau), g(\tau) ) + a^{2} \text{Var}_{\pi} (g(\tau)) \right], 
\end{equation}
which is potentially smaller than $\frac{1}{n} \text{Var}_{\pi}(f(\tau))$ if the covariance between $f$ and $g$ is sufficiently large. Under these conditions, we choose the value of $a$ that minimizes \autoref{eq:aaaaaa} \cite{weaver2013optimal}, namely, $a = \nicefrac{\text{Cov}_{\pi}(f(\tau), g(\tau))}{\text{Var}_{\pi}(g(\tau))}$. We then call the function $g$ a \emph{control variate} \cite{mcbook}. Also, although the quantities defining the best baseline $a$ are generally unavailable in closed form, one commonly uses a batch-based estimate of  $\text{Cov}_{\pi}(f(\tau), g(\tau))$ and $\text{Var}_{\pi}(g(\tau))$; the incurred bias is generally negligible relatively to the reduced variance \cite{ranganath14, shi2022gradient, roeder2017stl}. For vector-valued RVs, we let $a$ be a diagonal matrix and exhibit, in the next proposition, the optimal baseline minimizing the covariance matrix's trace. 

\begin{proposition}[Control variate for gradients] \label{prop:aaaa} 
    Let $f, g \colon \mathcal{P}_{\mathcal{S}} \rightarrow \mathbb{R}^{d}$ be vector-valued functions and $\pi$ be a probability measure on $\mathcal{P}_{\mathcal{S}}$. Consider a \emph{baseline} $a \in \mathbb{R}$ and assume $\mathbb{E}_{\pi}[g(\tau)] = 0$. Then, 
    \begin{equation*}
        \argmin_{a \in \mathbb{R}} \textrm{{\normalfont Tr }} \text{{\normalfont Cov}}_{\pi}[f(\tau) - a \cdot g(\tau)] \!\! = \!\! \frac{\mathbb{E}_{\pi}[g(\tau)^{T} (f(\tau) - \mathbb{E}_{\pi} [ f(\tau') ])]}{\mathbb{E}_{\pi}[g(\tau)^{T} g(\tau)]}.    
    \end{equation*}
\end{proposition}

Note that, when implementing the REINFORCE gradient estimator, the expectation we wish to estimate may be generally written as $\mathbb{E}_{P_{F}(s_{o}, \cdot)} \left[ \nabla_{\theta} f(\tau) + f(\tau) \nabla_{\theta} \log p_{F_\theta}(\tau)\right]$. For the second term, we use a leave-one-out estimator \cite{shi2022gradient}; see below. For the first term, we use $\nabla_{\theta} \log p_{F_\theta}$ as a control variate, which satisfies $\mathbb{E}_{P_{F}(s_{o}, \cdot)} \left[\nabla_{\theta} \log p_{F_\theta}(\tau | s_{o} ; \theta)\right] = 0$. Importantly, estimating the optimal baseline $a^{\star}$ in \autoref{prop:aaaa} cannot be done efficiently due to the non-linear dependence of the corresponding Monte Carlo estimator on the sample-level gradients \cite{baydin2018automatic}; i.e., it cannot be represented as a vector-Jacobian product, which is efficient to compute in reverse-mode automatic differentiation (\textit{autodiff}) frameworks \cite{paszke2019pytorch, jax2018github}. Consequently, we consider a linear approximation of both numerator and denominator defining $a^{\star}$ in \autoref{prop:aaaa}, which may be interpreted as an instantiation of the delta method \cite[Sec. 7.1.3]{schervish2012theory}. Then, given a batch $\{\tau_{1}, \dots, \tau_{N}\}$ of trajectories, we instead use  
\begin{equation} \label{eq:aaa} 
    \hat{a} = \frac{\left\langle \sum_{n=1}^{N} \nabla_{\theta} \log p_{F_\theta}(\tau_{n}), \sum_{n=1}^{N} \nabla_{\theta} f(\tau_{n})\right\rangle}{\epsilon + \left\|  \sum_{n=1}^{N} \nabla_{\theta} \log p_{F_\theta}(\tau_{n}) \right\|^{2}} 
\end{equation}
as the REINFORCE batch-based estimated baseline; $\langle \cdot, \cdot \rangle$ represents the inner product between vectors. Intuitively, the numerator is roughly a linear approximation to the covariance between $\nabla_{\theta} \log p_{F_\theta}$ and $\nabla_{\theta} f$ under $P_{F}$. In contrast, the denominator approximately measures the variance of $\nabla_{\theta} \log p_{F_\theta}$, and $\epsilon > 0$ is included to enhance numerical stability. As a consequence, for the reverse KL divergence, $\nabla_{\theta} f(\tau) = \nabla_{\theta} \log p_{F_\theta}(\tau)$, $\hat{a} \approx 1$ and the term corresponding to the expectation of $\nabla_{\theta} f(\tau)$ vanishes. We empirically find that this approach frequently reduces the variance of the estimated gradients by a large margin (see \autoref{fig:cv} above and \autoref{sec:e} below).    

\pp{Leave-one-out estimator.}\label{p:a} We now focus on obtaining a low-variance estimate of $\mathbb{E}_{\tau \sim P_{F}(s_{o}, \cdot)}[f(\tau) \nabla_{\theta} \log p_{F_\theta}(\tau)]$. As an alternative to the estimator of \autoref{prop:aaaa}, \citet{shi2022gradient} and \citet{salimans2014using} proposed a sample-dependent baseline of the form 
$    a(\tau_{i}) = \frac{1}{N - 1} \sum_{1 \le n \le N, n \neq i} f(\tau_{n})  $
for $i \in \{1, \dots, N\}$. The resulting estimator, 
\begin{equation*}
    \delta = \frac{1}{N} \sum_{n=1}^{N} \left( f(\tau_{n}) - \frac{1}{N - 1} \sum_{j=1, j \neq i}^{N} f(\tau_{j}) \right) \nabla_{\theta} \log p_{F_\theta}(\tau_{n}), 
\end{equation*}
is unbiased for $\mathbb{E}\left[ f(\tau) \nabla_{\theta} \log p_{F_\theta}(\tau) \right]$ due to the independence between $\tau_{i}$ and $\tau_{j}$ for $i \neq j$. Strikingly, $\delta$ can be swiftly computed with \textit{autodiff}: if $\mathbf{f} = (f(\tau_{n}))_{n=1}^{N}$ and $\mathbf{p} = \left(\log p_{F_\theta}(\tau_{n})\right)_{n=1}^{N}$, then 
\begin{equation}
    \delta = \nabla_{\theta} \frac{1}{N} \left\langle \text{sg} \left( \mathbf{f} - \frac{1}{N - 1} (\mathbf{1} - \mathbf{I}) \mathbf{f} \right), \mathbf{p} \right\rangle, 
\end{equation}
with $\text{sg}$ corresponding to a stop-gradient operation (e.g., represented by \texttt{lax.stop\_gradient} in JAX \cite{jax2018github} and \texttt{torch.detach} in PyTorch \cite{paszke2019pytorch}). Importantly, these techniques incur a minimal computational overhead to the stochastic optimization algorithms relative to the considerable reduction in variance they enact.

\noindent \textbf{Relationship with previous works.} Importantly,  \citet{malkin2023gflownets} used $\hat{a} = \frac{1}{N} \sum_{n} f(\tau_{n})$ as baseline and an importance-weighted aggregation to adjust for the off-policy sampling of trajectories, introducing bias in the gradient estimates and relinquishing guarantees of the optimization procedure. A learnable baseline independently trained to match $\hat{b}$ was also considered. This potentially entailed the inaccurate conclusion that the TB and DB are superior to standard divergence-based objectives. Indeed, the following section underlines that such divergence measures are sound and practical learning objectives for GFlowNets for a range of tasks.

\pp{Illustration of the control variates' effectiveness.} We train the GFlowNets using increasingly larger batches of $\{2^{i} \colon i \in [[5, 10]]\}$ trajectories with and without CVs. In this setting, \autoref{fig:cv} showcases the drastic reduction in the variance, represented by the covariance matrix's trace, of the estimated learning objectives' gradients w.r.t. the model's parameters promoted by the CVs. Impressively, as we show in \autoref{fig:grad}, this approach significantly increases the efficiency of the underlying stochastic optimization algorithm. See  \autoref{sec:e} and \autoref{app:additional} for further details. 


\section{Training GFlowNets with divergence measures} \label{sec:e}  

Our experiments serve two purposes. Firstly, we show in \autoref{sec:conv} that minimizing divergence-based learning objectives leads to competitive and often better approximations than the alternatives based on log-squared violations of the flow network's balance, offering a wider diversity of evaluation settings complementing prior evidence \cite{DBLP:journals/tmlr/ZimmermannLMN23, malkin2023gflownets}. Secondly, we highlight in \autoref{sec:red} that the reduction of variance enacted by the introduction of adequately designed control variates critically accelerates the convergence of GFlowNets. We consider widely adopted benchmark tasks from GFlowNet literature, described in \autoref{sec:env}, contemplating both discrete and continuous target distributions. For further information concerning the tasks' and models' hyperparameters, see \autoref{sec:app:e} and \autoref{app:details}. 

\subsection{Generative tasks} \label{sec:env} 

Below, we provide a high-level characterization of the generative tasks used for synthetic data generation and training. For a more rigorous description of \autoref{sec:p}, see \autoref{sec:app:e}.    

\noindent\textbf{Set generation} \cite{Bengio2021, pan2023better, pan2023generative, jang2024learning}\textbf{.} A state $s$ corresponds to a set of size up to a given $S$ and the terminal states $\mathcal{X}$ are sets of size $S$; a transition corresponds to adding an element from a deposit $\mathcal{D}$ to $s$. The IGP starts at an empty set, and the log-reward of a $x \in \mathcal{X}$ is $\sum_{d \in x} f(d)$ for a fixed $f \colon \mathcal{D} \rightarrow \mathbb{R}$.      

\noindent\textbf{Autoregressive sequence generation} \cite{sequence, malkin2022trajectory}\textbf{.} Similarly, a state is a seq. $s$ of max size $S$ and a terminal state is a seq. ended by an end-of-sequence token; a transition appends $d \in \mathcal{D}$ to $s$. The IGP starts with an empty sequence and, for $x \in \mathcal{X}$, $\log r(x) = \sum_{i=1\ldots |x|} g(i) f(x_{i})$ for functions $f, g$. 

\noindent\textbf{Bayesian phylogenetic inference (BPI)} \cite{zhou2024phylogfn}\textbf{.} A state $s$ is a forest composed of binary trees with labeled leaves and unlabelled internal nodes, and a transition amounts to joining the roots of two trees to a newly added node. Then, $s$ is terminal when it is a single connected tree --- called a \emph{phylogenetic tree}. Finally, given a dataset of nucleotide sequences, the reward function is the unnormalized posterior over trees induced by J\&C69's mutation model \cite{Jukes1969} and a uniform prior.   

\noindent\textbf{Mixture of Gaussians (GMs)} \cite{theory, zhang2023diffusion}\textbf{.} The IGP starts at $\mathbf{0} \in \mathbb{R}^{d}$ and proceeds by sequentially substituting each coordinate with a sample from a real-valued distribution. For a $K$-component GM, the reward of $\mathbf{x} \in \mathbb{R}^{d}$ is defined as $\sum_{k=} \alpha_{k} \mathcal{N}(\mathbf{x} | \mu_{k}, \Sigma_{k})$ with $\alpha_{k} \ge 0$ and $\sum_{k} \alpha_{k} = 1$. 

\noindent\textbf{Banana-shaped distribution}~\cite{rhodes2019variational, Mesquita2019}\textbf{.}  We use the same IGP implemented for a bi-dimensional GM. For $\mathbf{z} \in \mathbb{R}^{2}$, we set $r(\mathbf{x})$ to a normal likelihood defined on a quadratic function of $\mathbf{x}$, see \autoref{eq:banana} in the supplement. We use HMC samples as ground truth to gauge performance on this task. 


\begin{figure*}
    \center 
    \includegraphics[width=.9\textwidth]{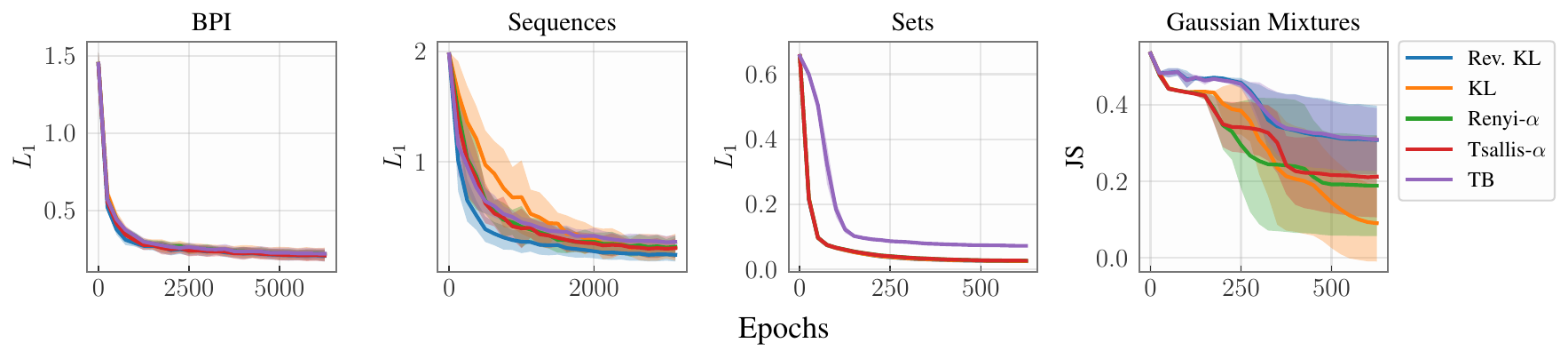} 
    \vspace{-4pt}
    \caption{\textbf{Divergence-based learning objectives often lead to faster training than TB loss.} Notably, contrasting with the experiments of \cite{malkin2023gflownets}, there is no single best loss function always conducting to the fastest convergence rate, and minimizing well-known divergence measures is often on par with or better than minimizing the TB loss in terms of convergence speed. Results were averaged across three different seeds. Also, we fix $\alpha = 0.5$ for both Tsallis-$\alpha$ and Renyi-$\alpha$ divergences.} 
    \label{fig:conv} 
    \vspace{-8pt}  
\end{figure*}

\begin{figure*}
    \centering
    \includegraphics[width=.8\textwidth]{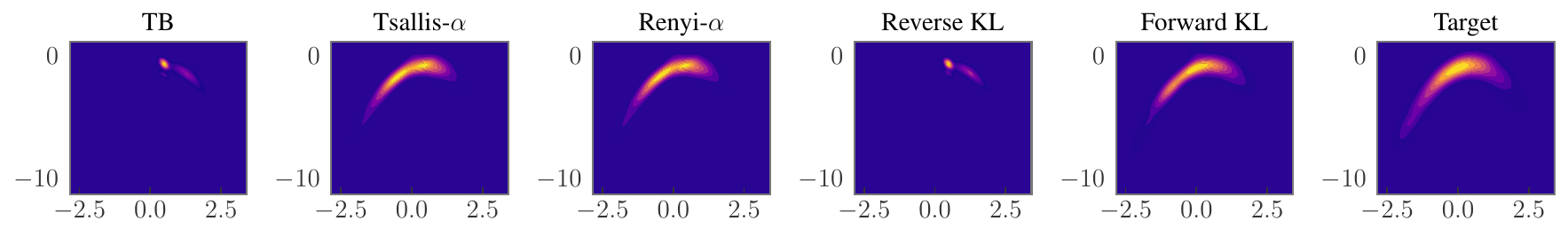}
        \vspace{-4pt}
    \caption{\textbf{Learned distributions for the banana-shaped target.} Tsallis-$\alpha$, Renyi-$\alpha$ and for. KL leads to a better model than TB and Rev. KL, which behave similarly --- as predicted by \autoref{prop:aaa}.}
    \vspace{-12pt} 
    \label{fig:banana}
\end{figure*}


\subsection{Assessing convergence speed} \label{sec:conv}

Next, we provide evidence that minimizing divergence-based objectives frequently leads to faster convergence than minimizing the standard TB loss \cite{malkin2022trajectory}. 

\noindent\textbf{Experimental setup.} We compare the convergence speed in terms of the rate of decrease of a measure of distributional error when using different learning objectives for a GFlowNet trained to sample from each of the distributions described in \autoref{sec:env}. For discrete distributions, we adopt the evaluation protocols of previous works \cite{Madan2022LearningGF, Bengio2021, malkin2022trajectory, pan2023generative} and compute the $L_{1}$ distance between the learned $p_{T}(x ; \theta)$ and target $r(x)$, namely, $\sum_{x \in \mathcal{X}} |p_{T}(x ;\theta) - \nicefrac{r(x)}{Z}|$. To approximate $p_{T}$, we use a Monte Carlo estimate of  
    $p_{T}(x ; \theta) = \mathbb{E}_{\tau \sim P_{B}(x, \cdot)} \left[ \nicefrac {p_{F_\theta}(\tau | s_{o} ; \theta)}{p_{B}(\tau | x)} \right]$.   
For continuous distributions, we echo \cite{theory, zhang2023diffusion} and compute Jensen-Shannon's divergence between $P_{T}(x ; \theta)$ and $R(x)$: 
\begin{equation*}    
    \mathcal{D}_{JS} [ P_{T} || R ] = \nicefrac{1}{2} \left( \mathcal{D}_{KL} [P_{T} || M ] + \mathcal{D}_{KL} [ R || M ] \right) 
    = \mathbb{E}_{x \sim P_{T}} \left[ \log \nicefrac{p_{T}(x)}{m(x)} \right] + \mathbb{E}_{x \sim R} \left[ \log \nicefrac{r(x)}{Z m(x)} \right],  
\end{equation*}
with $M(B) = \nicefrac{1}{2} \left( P_{T}(B) + \nicefrac{R(B)}{R(\mathcal{X})} \right)$ being the averaged measure of $P_{T}$ and $R$ and $m$ its corresponding density relatively to the reference measure $\mu$. Remarkably, in the context of approximating mixture of Gaussian distributions, we can directly sample from the target to estimate $\mathcal{D}_{KL}[ R || M ]$, and the autoregressive nature of the generative process ensures that $p_{T}(x) = p_{F_\theta}(\tau | s_{o})$ for the unique trajectory $\tau$ starting at $s_{o}$ and finishing at $x$. Hence, we get an unbiased estimate of $\mathcal{D}_{KL}[P_{T} || M]$. 


\captionsetup[figure]{font=small}
\begin{wraptable}{!h}{.5\textwidth}  
    \centering
        \caption{Divergence minimization achieves better than or similar accuracy compared to enforcing TB.} 
    \resizebox{\linewidth}{!}{
    \begin{tabular}{c|c|c|c|c}
         & BPI & Sequences & Sets & GMs \\   
         \hline 
         TB & $0.22 \textcolor{gray}{\scriptstyle \pm 0.04}$ & $0.28 \textcolor{gray}{\scriptstyle \pm 0.06}$ & $0.07 \textcolor{gray}{\scriptstyle \pm 0.00}$ & $0.31 \textcolor{gray}{\scriptstyle \pm 0.08}$  \\ 
         Rev. KL & $0.21 \textcolor{gray}{\scriptstyle \pm 0.04}$ & $\textbf{0.16} \textcolor{gray}{\scriptstyle \pm 0.06}$ & $\textbf{0.03} \textcolor{gray}{\scriptstyle \pm 0.00}$ & $0.31 \textcolor{gray}{\scriptstyle \pm 0.09}$ \\
         For. KL & $0.22 \textcolor{gray}{\scriptstyle \pm 0.04}$ & $0.23 \textcolor{gray}{\scriptstyle \pm 0.12}$ & $\textbf{0.03} \textcolor{gray}{\scriptstyle \pm 0.00}$ & $\textbf{0.09} \textcolor{gray}{\scriptstyle \pm 0.10}$ \\  
         Renyi-$\alpha$ & $0.22 \textcolor{gray}{\scriptstyle \pm 0.03}$ & $0.23 \textcolor{gray}{\scriptstyle \pm 0.10}$ & $\textbf{0.03} \textcolor{gray}{\scriptstyle \pm 0.00}$ & $0.19 \textcolor{gray}{\scriptstyle \pm 0.13}$ \\ 
         Tsallis-$\alpha$ & $0.21 \textcolor{gray}{\scriptstyle \pm 0.04}$ & $0.22 \textcolor{gray}{\scriptstyle \pm 0.09}$ & $\textbf{0.03} \textcolor{gray}{\scriptstyle \pm 0.00}$ & $0.21 \textcolor{gray}{\scriptstyle \pm 0.11}$  
    \end{tabular}
    }
    \label{tab:conv}
\end{wraptable}
\captionsetup[figure]{font=normal}
\pp{Results.} \autoref{fig:conv} shows that the procedure minimizing divergence-based measures accelerates the training convergence of GFlowNets, whereas \autoref{fig:banana} (for the banana-shaped distribution) and \autoref{tab:conv} highlight that we obtain a more accurate model with a fix compute budget. The difference between learning objectives is not statistically significant for the BPI task. Also, we may attribute the superior performance of reverse KL compared to the forward in the sequence generation task to the high variance of the importance-sampling-based gradient estimates. Indeed, the observed differences disappear when we increase the batch of trajectories to reduce the estimator's variance (see \autoref{fig:seqs} in \autoref{app:additional}). In conclusion, our empirical results based on experiments testing diverse generative settings and expanding prior art \cite{DBLP:journals/tmlr/ZimmermannLMN23, malkin2023gflownets, theory}, shows that training methods based on minimizing \textit{f}-divergence VI objectives with adequate CVs implemented are practical and effective in many tasks.

\subsection{Reducing the variance of the estimated gradients} \label{sec:red}

\autoref{fig:cv} demonstrates that implementing CVs for the REINFORCE estimator reduces the noise level of gradient estimates significantly. This reduction in variance also accelerates training convergence. To illustrate this, we use the same experimental setup from \autoref{sec:conv} and analyze the learning curves for each divergence measure with and without control variates.



\begin{figure}[H]
    \centering
    \includegraphics[width=.9\textwidth]{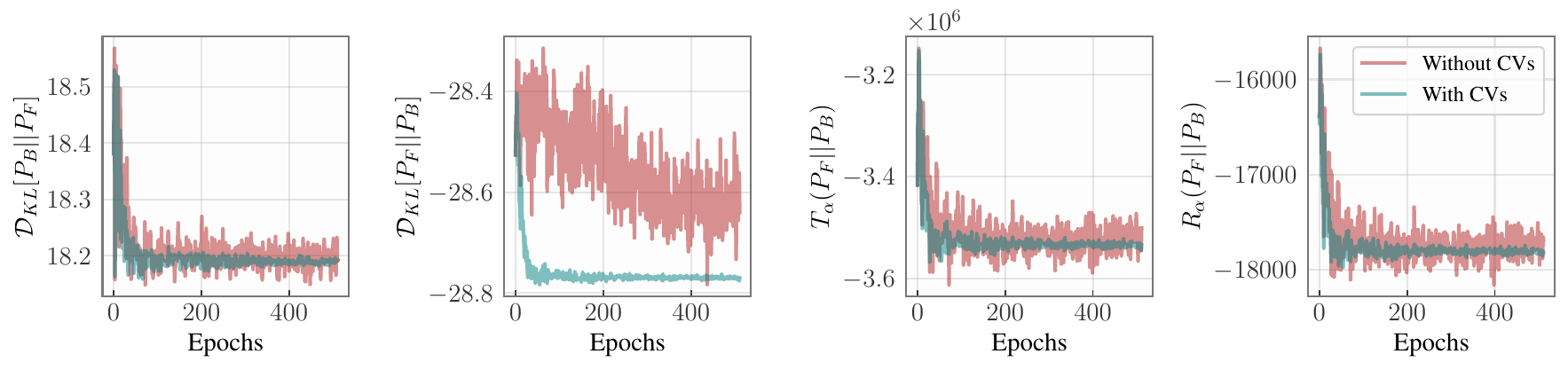}
    \caption{\textbf{Learning curves for different objective functions} in the task of set generation. The reduced variance of the gradient estimates notably increases training stability and speed.}
    \vspace{-12pt} 
    \label{fig:grad}
\end{figure}

\noindent\textbf{Results.} \autoref{fig:grad} shows that the implemented gradient reduction techniques significantly enhance the learning stability of GFlowNets and drastically accelerate training convergence when minimizing the reverse KL divergence. Our results indicate that well-designed CVs for gradient estimation can greatly benefit GFlowNets training. Notably, researchers have observed similar improvements using reduced-variance gradient estimators in Langevin dynamics simulations \cite{huang21ld, kumar16ld, kinoshita2022ld} and policy gradient methods for RL \cite{xu20pg, papini18pg}.

\section{Conclusions, limitations and broader impact} \label{sec:s} 


We showed in a comprehensive range of experiments that \textit{f}-divergence measures commonly employed in VI --- forward KL, reverse KL, Renyi-$\alpha$, and Tsallis-$\alpha$ --- are effective learning objectives for training GFlowNets, achieving competitive performance to flow-based training objectives. Variance reduction for the gradient estimators of the \textit{f}-divergences proved essential for competitive results, clarifying prior results regarding VI training \cite{DBLP:journals/tmlr/ZimmermannLMN23, malkin2022trajectory, theory} in the general setting of \textit{f}-divergences minimization.
Additionally, we developed a theoretical connection between GFlowNets and VI beyond the setup of finitely supported measures, establishing results for arbitrary topological spaces. 

While we observed promising results for $\alpha=0.5$, there might be different choices of $\alpha$ that, depending on the application, might strike a better explore-exploit tradeoff and incur faster convergence. Thus, thoroughly exploring different $\alpha$ might be especially useful to practitioners.

Overall, our work highlights the potential of the once-dismissed VI-inspired schemes for training GFNs, paving the way for further research towards improving the GFlowNets by drawing inspiration from the VI literature. For instance, one could develop $\chi$-divergence-based losses for GFNs \cite{dieng2017variational}, combine GFNs with MCMC using \citet{ruiz2019contrastive}'s divergence, or employ an objective similar to that of importance-weighted autoencoders \cite{burda2016importance}.  


\section*{Acknowledgements}

This work was supported by the Fundação Carlos Chagas Filho de Amparo à Pesquisa do Estado do Rio de Janeiro FAPERJ (SEI-260003/000709/2023), the São Paulo Research Foundation FAPESP (2023/00815-6), the Conselho Nacional de Desenvolvimento Científico e Tecnológico CNPq (404336/2023-0), and the Silicon Valley Community Foundation through the University Blockchain Research Initiative (Grant 2022-199610).

\bibliographystyle{abbrvnat} 
\bibliography{bibliography} 

\newpage 
\onecolumn 

\appendix 

\section{Related works} \label{sec:r} 

\noindent\textbf{Generative Flow Networks.} GFlowNets \cite{Bengio2021, Foundations} were initially proposed as a reinforcement learning algorithm for finding diverse high-valued states in a discrete environment by sampling from a distribution induced by a reward function. Shortly after, they were extended to sample from complex distributions in arbitrary topological spaces \cite{theory}. Remarkably, this family of models was successfully applied to problems as diverse as structure learning and causal discovery \cite{deleu2023joint, deleu2022bayesian, dasilva2023humanintheloop}, discrete probabilistic modeling and graphical models \cite{discretegfn_i, discretegfn_ii, discretegfn_iii, falet2023deltaai}, combinatorial optimization and stochastic control \cite{Zhang2023, zhang2023diffusion}, drug discovery \cite{Bengio2021, jain2023gflownets, nica2022evaluating}, design of biological sequences \cite{sequence}, natural language processing \cite{hu2023amortizing}, and aerial scene classification \cite{dutta2023towards}. Concomitantly to these advances, there is a growing literature concerned with the development of more efficient training algorithms for GFlowNets \cite{shen23gflownets, malkin2022trajectory, Foundations, kim2023local} --- primarily drawing inspiration from existing techniques in the reinforcement learning literature \cite{pan2023better, pan2023generative, mohammadpour2023maximum, tiapkin2024generative}. In the same spirit, \citet{tiapkin2024generative} showed it is possible to frame GFlowNets as an entropy-regularized reinforcement learning. In a study closely related to ours, \citet{malkin2023gflownets} proved the equivalence between GFlowNets and hierarchical variational inference (HVI) for discrete distributions; however, when training GFlowNets using divergence-based methods from the VI literature, the authors found no improvement relatively to the traditional flow-matching objectives. Thus, extending beyond discrete distributions, this work provides a definitive analysis of training GFlowNets by directly optimizing a set of divergences typically employed in variational inference training, given a clear context and conditions for effective use of divergence objectives for efficient learning procedures applied on GFlowNets models.  

\noindent\textbf{Reinforcement Learning as Inference.} Reinforcement Learning (RL) has been studied as a form of probabilistic inference extensively, generating relevant insights in the literature, and alternatively referred to as \textit{control as inference}. \citet{DBLP:conf/cdc/Todorov08} demonstrates a duality between estimation and optimal control, establishing conditions where estimation algorithms could applied for control problems. \citet{DBLP:journals/ml/KappenGO12-optimcinf} demonstrated that optimal control problems could be framed as inference problems in graphical models, providing a unified perspective for solving control tasks. \citet{DBLP:levine-rl-control-inf} presents a complete and modern RL formulation, linking with VI in particular. \citet{DBLP:conf/nips/RudnerPMGL21-outcome-drivenrlvi} integrates even further RL with VI methods, demonstrating the conceptual and algorithmic gains of leveraging outcome-driven RL with variational inference to optimize policy distributions. Developing further, \citet{DBLP:conf/icml/ToussaintS06} applies approximate probabilistic inference methods to solve Markov Decision Processes (MDPs) with discrete and continuous states. The approach also aligns with model-based RL techniques, such as \textit{PILCO}, which utilizes probabilistic models to enhance data efficiency in policy search \cite{DBLP:conf/icml/DeisenrothR11-model-based-policy}. Recent work by \citet{DBLP:journals/corr/abs-2402-10309} positions discrete probabilistic inference as a control problem in multi-path environments, highlighting the synergy between control theory and probabilistic modeling in the context of GFlowNets. This body of works relates to the approach presented in this paper, comparing optimization of trajectory balance and flow-matching losses related to sequential decisions modeled by the GFlowNet with \textit{f}-divergence measures minimization procedures -- related to approximated variational inference and generalized posterior inference \cite{DBLP:journals/jmlr/KnoblauchJD22, DBLP:conf/nips/WildHS22, li2016renyi, DBLP:journals/ml/JordanGJS99, DBLP:journals/ftml/WainwrightJ08}. 

\noindent\textbf{Divergence measures and gradient reduction for VI.} Approximate inference via variational inference (VI) methods \cite{DBLP:journals/ml/JordanGJS99, DBLP:journals/ftml/WainwrightJ08, bishop2007, Blei2017} initially relied on message passing and coordinate ascent methods to minimize the KL divergence of an unnormalized distribution and a proposal in a parameterized tractable family of distributions. Despite the initial generality of the optimization perspective, the concrete implementation of algorithms often requires specialized update equations and learning objectives for specific classes of models. On the other hand, the development of algorithms and software for automatic differentiation \cite{baydin2018automatic} and stochastic gradient estimators \cite{DBLP:journals/jmlr/MohamedRFM20} unlocked the potential application of generic gradient-based optimization algorithms in inference and learning tasks for a comprehensive class of models. Seminal works such as Black-Box VI (BBVI) \cite{ranganath14}, using the REINFORCE/score function estimator, and Automatic Differentiation VI (ADVI) \cite{DBLP:journals/jmlr/KucukelbirTRGB17}, using reparameterization and change-of-variables, demonstrated practical algorithms for Bayesian inference in generic models, including models combining classical statistical modeling with neural networks. Overall, \citet{DBLP:journals/jmlr/MohamedRFM20} explain the development of the main gradient estimators: the score function \cite{Williams1992, carbonetto2009stochastic, ranganath14, yin2018semi}, and the pathwise gradient estimator, also known as the parametrization trick \cite{rezende2015variational, kingma2014semi, kingma2023variational}. The vanilla REINFORCE/score function estimator has notoriously high variance \cite{Williams1992, richter2020vargrad, carbonetto2009stochastic, ranganath14}, which prompted a body of work exploring variance reduction techniques. In the original BBVI proposal, \citet{ranganath14} explored Rao-Blackwellization, combining iterated conditional expectations and control variates, using the score function estimator (given its zero expectation) as a control variate. Subsequent works have continued to refine these techniques; \citet{DBLP:conf/icml/LiuRTJM19} uses Rao-Blackwellized stochastic gradients for discrete distributions, while \citet{DBLP:conf/aistats/KimMG24} and \citet{DBLP:conf/aistats/WangGD24} explored joint control variates and provable linear convergence in BBVI. Additionally, \citet{DBLP:conf/icml/Domke20} and \citet{DBLP:conf/nips/Domke19} provided smoothness and gradient variance guarantees, further enhancing the robustness of score function estimator for VI methods. Our work demonstrates that effective variance reduction techniques applied to a \textit{f}-divergence minimization training can significantly enhance the convergence speed and stability of the procedure. In theory and practice, we observed high compatibility between our results of variance-reduced \textit{f}-divergence GFlowNets training and the body of work of variance-reduced score-function estimators for VI. Furthermore, by showing that these techniques apply to a broad class of models and optimization objectives, including continuous and mixed structured supports, we move GFlowNets' \textit{f}-divergence minimization training closer to recent notions of generalized Bayesian inference and generalized VI\cite{DBLP:journals/jmlr/KnoblauchJD22} and variational inference in function spaces \cite{DBLP:conf/nips/WildHS22} -- with the common thread of casting posterior inference as an optimization problem guided by some divergence measure. This generalization can enable applications of GFlowNets to a diverse range of machine learning tasks, enhancing their versatility and practical relevance.

\section{Detailed description of the generative tasks} \label{sec:app:e} 

\noindent\textbf{Set generation} \cite{Bengio2021, pan2023better, pan2023generative, jang2024learning}\textbf{.} $\mathcal{S}$ contains the sets of size up to $N$ with elements extracted from a fixed deposit $\mathcal{D}$ of size $D \ge N$ and $s_{o} = \emptyset$. For $s \in \mathcal{S}$ with $|s| < N$, $\kappa_{f}(s, \cdot)$ is a counting measure supported at (the $\sigma$-algebra induced by) $\{s \cup \{d\} \colon d \in \mathcal{D} \setminus s\}$; for $|s| = N$, $\kappa_{f}(s, \cdot) = \delta_{s_{f}}$. Then, $\mathcal{X} = \{s \in \mathcal{S} \colon |s| = N\}$. Similarly, $\kappa_{b}(s, \cdot)$'s support is $\{s \setminus \{d\} \colon d \in \mathcal{D}\}$ for $s \neq s_{o}$. We define the unnormalized target's density by $\log r(x) = \sum_{d \in x} f(d)$ for a fixed function $f \colon \mathcal{D} \rightarrow \mathbb{R}$. We parameterize $p_{F}(s, \cdot)$ as a deep set \cite{deepsets} and fix $p_{B}(s, \cdot)$ as a uniform density for $s \in \mathcal{S}$.   

\noindent\textbf{Autoregressive sequence generation} \cite{sequence, malkin2022trajectory}\textbf{. } A \emph{sequence} $s$ in $\mathcal{D}^{n}$, for any $K > n$, is bijectively associated to an element of $\mathcal{D} \times [K]$ by $s \mapsto \left\{ (s_{m}, m) \colon 1 \le m \le n\right\} \cup \left\{ (\bot, m) \colon K \ge m > n\right\}$; $\bot$ is a token denoting the sequence's end. In this context, we let $\mathcal{S} \subset \mathcal{P}(\mathcal{D} \times [N + 1])$ be the set of sequences of size up to $N$, i.e., if $s \in \mathcal{S}$ and $(\bot, n + 1) \in s$, then $(d, m) \in s$ iff $d = \bot$ for $n < m \le N + 1$ and there is $v \in (\mathcal{D} \cup \{\bot\})^{n}$ such that $(v_{m}, m) \in s$ for $m \le n$; the initial state is $s_{o} = \emptyset$. For conciseness, we write $d \notin s$, meaning that $(d, i) \notin s$ for every $i$. Next, $\kappa_{f}(s, \cdot)$ is the counting measure supported at $\{s \cup \{(d, |s| + 1)\} \colon d \in \mathcal{D} \cup \{\bot\}\}$ if $|s| < N$ and $\bot \notin s$; at $\{s \cup \{(\bot, N + 1)\}\}$ if $|s| = N$; and at $\{s_{f}\}$ otherwise. Thus, $\mathcal{X} = \{s \in \mathcal{S} \colon \bot \in s\}$. Also, $k_{b}(s, \cdot)$ is supported at $\{s \setminus \{(d, |s|)\} \colon d \in \mathcal{D}\}$, which has a single element corresponding to the removal of the element of $s$ of the largest index. On the other hand, the target's density is $\log r(x) = \sum_{(d, i) \in x \colon d \neq \bot} f(d) g(i)$ for functions $f, g \colon \mathcal{D} \rightarrow \mathbb{R}$. We employ an MLP to parameterize $p_{F}(s, \cdot)$. 

\noindent\textbf{Bayesian phylogenetic inference (BPI)} \cite{zhou2024phylogfn}\textbf{.} A (rooted) \emph{phylogeny} is a complete binary tree with labeled leaves and weighted edges; each leaf corresponds to a biological species, and the edges' weights are a measurement of evolutionary time. Formally, we let $\mathcal{B}$ be the set of observed biological species and $\mathcal{V}_{\mathcal{B}}$ be the set of $|\mathcal{B}| + 1$ unobserved species. 
Next, we represent a phylogeny over $\mathcal{B}$ as a weighted directed tree $G_{\mathcal{B}} = (\mathcal{B} \cup \mathcal{V}_{\mathcal{B}}, E_{\mathcal{B}}, \omega_{\mathcal{B}})$ with edges $E_{\mathcal{B}}$ featured with a weight assignment $\omega_{\mathcal{B}}$; we denote its root by  $r(G_{\mathcal{B}})$. 
Under these conditions, we define $\mathcal{S} = \left\{ \bigcup_{k=1}^{K} G_{\mathcal{F}_{k}} \colon \bigsqcup_{k} \mathcal{F}_{k} = \mathcal{B} \land \mathcal{G}_{\mathcal{F}_{k}} \text{ is a tree} \right\}$ as the set of forests built upon phylogenetic trees over disjoint subsets of $\mathcal{B}$; $\bigsqcup$ represents a disjoint union of non-empty sets and $s_{o} = \bigcup_{b \in \mathcal{B}} G_{\{b\}}$ is the forest composed of singleton trees $G_{\{b\}}$. 
Also, we define the \emph{amalgamation} of phylogenies $G_{\mathcal{F}_{k}}$ and $G_{\mathcal{F}_{j}}$, $\mathcal{A}(G_{\mathcal{F}_{k}}, G_{\mathcal{F}_{j}})$, as the tree obtained by joining their roots $r(G_{\mathcal{F}_{k}})$ and $r(G_{\mathcal{F}_{j}})$ to a new node $r(G_{\mathcal{F}_{k}} \cup G_{\mathcal{F}_{j}})$, with a corresponding extension of the edges' weights. In contrast, the \emph{dissolution} of a tree $G_{\mathcal{F}}$, $\mathcal{R}(G_{\mathcal{F}})$, returns the union of the two subtrees obtained by removing $r(G_{\mathcal{F}})$ from $G_{\mathcal{F}}$.  
Then, $\kappa_{f}(s, \cdot)$ is the counting measure supported at $\left\{ \bigcup_{k=1, k \neq i, j}^{K} G_{\mathcal{F}_{k}} \cup \mathcal{A}(G_{\mathcal{F}_{i}}, G_{\mathcal{F}_{j}}) \colon (i, j) \in [K]^{2}, \, i \neq j \right\}$ with $s = \bigcup_{k=1}^{K} G_{\mathcal{F}_{k}}$ and $K \ge 2$; if $s = \mathcal{G}_{\mathcal{B}}$, $\kappa_{f}(s, \cdot) = \delta_{s_{f}}$. Hence, $\mathcal{X}$ is the set of phylogenies over $\mathcal{B}$. Likewise, $\kappa_{b}(s, \cdot)$'s support is $\left\{ \bigcup_{k=1, k \neq j}^{K} G_{\mathcal{F}_{k}} \cup \mathcal{R}(G_{\mathcal{F}_{i}}) \colon i \in [K] \land r(G_{\mathcal{F}_{i}}) \notin \mathcal{B} \right\}$ for $s = \bigcup_{k=1}^{K} G_{\mathcal{F}_{k}}$ and $K \le |\mathcal{B}|$. Finally, the unnormalized target is the posterior distribution defined by JC69's mutation model \cite{Jukes1969} given a data set of genome sequences of the species in $\mathcal{B}$. More specifically, we let the prior be a uniform distribution and compute the model-induced likelihood function by the efficient Felsenstein's algorithm \cite{Felsenstein1981}. We assume the edges' weights are constant. See \cite{Yang2014} for further details. We parameterize $p_{F}(s, \cdot)$ with GIN \cite{xu2018powerful} and fix $p_{B}(s, \cdot)$ as an uniform distribution. 

\noindent\textbf{Mixture of Gaussians} \cite{theory, zhang2023diffusion}\textbf{.} The training of GFlowNets in continuous spaces is challenging, and the problem of designing highly expressive models in this setting is still unaddressed \cite{deleu2023joint, theory}. However, as we show in \Cref{sec:conv}, divergence-based measures seem to be very effective learning objectives for autoregressive sampling of a sparse mixture of Gaussians. For a $d$-dimensional Gaussian distribution, $\mathcal{S} = \{\{(0, 0), (x_{i}, i) \colon 1 \le i \le n\}, \colon n \le d, \, x \in \mathbb{R}^{n}\} \subset \{(0, 0)\} \cup \mathcal{P}(\mathbb{R} \times [d])$ and $s_{o} = (0, 0)$; note $\mathcal{S}$ is isomorphic to $\mathbb{R}^{d}$. Also, for $s = \{(x_{i}, i)\}_{i=1}^{n}$, $\kappa_{f}(s, \cdot)$ is Lebesgue's measure at $\left\{ s \cup (x, n + 1) \colon x \in \mathbb{R} \right\}$ if $n < d$ and $\kappa_{f}(s, \cdot) = \delta_{s_{f}}$ otherwise. In particular, $\mathcal{X} = \{s \in \mathcal{S} \colon \max_{(x, i) \in s} i = d\}$. Moreover, $\kappa_{b}(s, \cdot)$ is a measure on $\{ s \setminus (x, |s|) \colon x \in \mathbb{R}\}$, which is isomorphic to $\mathbb{R}$, which is a singleton. We define the target's density with a homogeneous mixture of Gaussian distributions, $\frac{1}{K} \sum_{i=1}^{K} \mathcal{N}(\mu_{i}, \sigma^{2} I)$ with $\mu_{i} \in \mathbb{R}^{d}$. We similarly define $P_{F}(s, \cdot)$ as a mixture of one-dimensional Gaussians with mean and variance learned via an MLP \cite{theory}. 

\noindent\textbf{Banana distribution.} \cite{rhodes2019variational, Mesquita2019} We consider sampling from the banana distribution, defined by 
\begin{equation} \label{eq:banana} 
 \mathcal{N}\left( \begin{bmatrix} x_{1} \\ x_{2} + x_{1}^{2} + 1 \end{bmatrix} \bigg| \begin{bmatrix} 0 \\ 0 \end{bmatrix}, \begin{bmatrix} 1 & 0.9 \\ 0.9 & 1 \end{bmatrix}\right). 
\end{equation} 
Given its geometry and shape, this distribution is a common baseline in the approximate Bayesian inference literature \cite{zhang2017hamiltonian, yu2023semi, tran2016adaptive}. This task is identical to sampling from a mixture of Gaussian distributions, except for the different target density specified by the model in \Cref{eq:banana}. Also, we rely on the implementation Hamiltonian Monte Carlo (HMC) \cite{neal2011mcmc, betancourt2017conceptual} provided by Stan \cite{carpenter2017stan} to obtain accurate samples from \eqref{eq:banana}.  

\section{Proofs}\label{app:proofs}

We will consider the measurable space of \emph{trajectories} $(\mathcal{P}_{\mathcal{S}}, \Sigma_{P})$, with $\mathcal{P}_{\mathcal{S}} = \{ (s, s_{1}, \dots, s_{n}, s_{f}) \in \mathcal{S}^{n + 1} \times \{s_{f}\} \colon 0 \le n \le N - 1 \}$ and $\Sigma_{P}$ as the $\sigma$-algebra generated by $\bigcup_{n=1}^{N + 1} \Sigma^{\otimes n}$. For notational convenience, we use the same letters for representing the measures and kernels of $(\mathcal{S}, \Sigma)$ and their natural product counterparts in $(\mathcal{P}_{\mathcal{S}}, \Sigma_{P})$, which exist by Carathéodory extension's theorem \cite{david}; for example, $\nu(B) = \nu^{\otimes n}(B)$ for $B = (B_{1}, \dots, B_{n}) \in \Sigma^{\otimes n}$ and $p_{F_\theta}(\tau | s_{o} ; \theta)$ is the density of $P_{F}^{\otimes n + 1}(s_{o}, \cdot)$ for $\tau = (s_{o}, s_{1}, \dots, s_{n}, s_{f})$ relatively to $\mu^{\otimes n}$. In this case, we will write $\tau$ for a generic element of $\mathcal{P}_{\mathcal{S}}$ and $x$ for its terminal state (which is unique by \autoref{def:aaa}).  

\subsection{Proof of \autoref{prop:aaa}} 

We will show that the gradient of the expected on-policy TB loss matches the gradient of the KL divergence between the forward and backward policies. Firstly, note that 
\begin{equation*}
    \begin{aligned}
        \nabla_{\theta} \mathcal{D}_{KL} [ P_{F} || P_{B} ] &= \nabla_{\theta} \mathbb{E}_{\tau \sim P_{F}(s_{o}, \cdot)} \left[ \log \frac{p_{F}(\tau | s_{o} ; \theta)}{p_{B}(\tau)} \right] \\ 
        &\hspace{-12pt}= \nabla_{\theta} \int_{\tau} \log \frac{p_{F}(\tau | s_{o} ; \theta)}{p_{B}(\tau)} \mathrm{d}P_{F}(s_{o}, \mathrm{d}\tau) \\ 
        &\hspace{-12pt}= \nabla_{\theta} \int_{\tau} \log \frac{p_{F}(\tau | s_{o} ; \theta)}{p_{B}(\tau)} p_{F}(\tau | s_{o} ; \theta) \mathrm{d}\kappa_{f}(s_{o}, \mathrm{d}\tau) \\ 
        &\hspace{-12pt}= \int_{\tau} \nabla_{\theta} \log \frac{p_{F}(\tau | s_{o} ; \theta)}{p_{B}(\tau)} P_{F}(s_{o}, \mathrm{d}\tau) \\ 
        &\hspace{-12pt}+ \int_{\tau} \log \frac{p_{F}(\tau | s_{o} ; \theta)}{p_{B}(\tau)} \nabla_{\theta} p_{F}(\tau | s_{o} ; \theta) \mathrm{d}\kappa_{f}(s_{o}, \mathrm{d}\tau) 
    \end{aligned}
\end{equation*}
by Leibniz's rule for integrals and the product rule for derivatives. Then, since $\nabla_{\theta} f(\theta) = f(\theta) \nabla \log f(\theta)$ for any differentiable function $f \colon \theta \mapsto f(\theta)$, 
\begin{equation} \label{eq:ww} 
    \begin{aligned} 
        \nabla_{\theta} \mathcal{D}_{KL} \left[ P_{F} || P_{B} \right] & \\ 
        = \!\!\!\!\! \underset{\tau \sim P_{F}(s_{o}, \cdot)}{\mathbb{E}} \left[ \nabla_{\theta} \log p_{F}(\tau | s_{o}) + \log \frac{p_{F}(\tau | s_{o})}{p_{B}(\tau)}  \nabla_{\theta} \log p_{F}(\tau | s_{o}) \right] & \\
        = \!\!\!\!\! \underset{\tau \sim P_{F}(s_{o}, \cdot)}{\mathbb{E}} \left[ \log \frac{p_{F}(\tau | s_{o})}{p_{B}(\tau)}  \nabla_{\theta} \log p_{F}(\tau | s_{o}) \right]; 
    \end{aligned} 
\end{equation}
we omitted the dependency of $P_{F}$ (and of $p_{F}$ thereof) on the parameters $\theta$ for conciseness. On the other hand, 
\begin{equation}
    \nabla_{\theta} \mathcal{L}_{TB}(\tau ; \theta) = 2 \left( \log \frac{p_{F}(\tau | s_{o} ; \theta)}{p_{B}(\tau)} \right) \nabla_{\theta} \log p_{F}(\tau)  
\end{equation}
by the chain rule for derivatives. Thus, 
\begin{equation}
    \mathbb{E}_{\tau \sim P_{F}(s_{o}, \cdot)} \nabla_{\theta} \mathcal{L}_{TB}(\tau ; \theta) = 2 \nabla_{\theta} \mathcal{D}_{KL} [ P_{F} || P_{B} ], 
\end{equation}
ensuring that the equivalence between $\mathcal{L}_{TB}$ and $\mathcal{D}_{KL}$ in terms of expected gradients holds in a context broader than that of finitely supported distributions  \cite{malkin2023gflownets}.  

\subsection{Proof of \autoref{lemma:a}} 

Henceforth, we will recurrently refer to the score estimator for gradients of expectations \cite{Williams1992}, namely, 
\begin{equation} \label{eq:w} 
    \begin{split} 
       \nabla_{\theta} \underset{\tau \sim P_{F}(s_{o}, \cdot)}{\mathbb{E}} \left[ f_{\theta}(\tau) \right] = 
        \underset{\tau \sim P_{F}(s_{o}, \cdot)}{\mathbb{E}} \left[ \nabla_{\theta} f_{\theta}(\tau) + f_{\theta}(\tau) \nabla_{\theta} \log p_{F}(\tau | s_{o} ; \theta) \right], 
   \end{split} 
\end{equation}
which can be derived using the arguments of the preceding section. In this context, the Renyi-$\alpha$'s divergence satisfies 
\begin{equation*}
    \nabla_{\theta} R_{\alpha} ( P_{F} || P_{B} ) = \frac{\nabla_{\theta} \mathbb{E}_{\tau \sim P_{F}(s_{o}, \cdot)} [ g(\tau, \theta) ]}{(\alpha - 1) \mathbb{E}_{\tau \sim P_{F}(s_{o}, \cdot)} g(\tau, \theta) }, 
\end{equation*}
with $g(\tau ; \theta) = \left( \nicefrac{p_{B}(\tau | x) r(x)}{p_{F}(\tau | s_{o} ; \theta)} \right)^{1 - \alpha}$ and $\alpha \neq 1$;  similarly, the Tsallis-$\alpha$'s divergence abides by 
\begin{equation}
    \nabla_{\theta} T_{\alpha} ( P_{F} || P_{B} ) = \frac{1}{(\alpha - 1)} \nabla_{\theta} \mathbb{E}_{\tau \sim P_{F}(s_{o}, \cdot)}  [ g(\tau, \theta ) ]. 
\end{equation} 
The statement then follows by substituting $\nabla_{\theta} \mathbb{E}_{\tau \sim P_{F}(s_{o}, \cdot)} [ g(\tau, \theta) ]$ with the corresponding score estimator given by \Cref{eq:w}.  

\subsection{Proof of \autoref{lemma:aa}} 

\noindent\textbf{Forward KL divergence.} The gradient of $\mathcal{D}_{KL}[ P_{B} || P_{F} ]$ is straightforwardly obtained through the application of Leibniz's rule for integrals, 
\begin{equation*}
    \nabla_{\theta} \mathcal{D}_{KL}[P_{B} || P_{F}] = - \mathbb{E}_{\tau \sim P_{B}(s_{f}, \cdot)} \left[ \nabla_{\theta} \log p_{F}(\tau | s_{o} ; \theta) \right],  
\end{equation*}
since the averaging distribution $P_{B}$ do not depend on the varying parameters $\theta$. However, as we compute Monte Carlo averages over samples of $P_{F}$, we apply an importance reweighting scheme \cite[Chapter 9]{mcbook} to the previous expectation to infer that, up to a positive multiplicative constant, 
\begin{equation*}
    \begin{split} 
        \nabla_{\theta} \mathcal{D}_{KL}[ P_{B} || P_{F} ] 
        \stackrel{C}{=} - \mathbb{E}_{\tau \sim P_{F}(s_{o}, \cdot)} \left[ \frac{p_{B}(\tau | x) r(x)}{p_{F}(\tau | s_{o} ; \theta)} \nabla_{\theta} \log p_{F}(\tau | s_{o} ; \theta) \right], 
    \end{split} 
\end{equation*}
with $\stackrel{C}{=}$ denoting equality up to a positive multiplicative constant. We emphasize that most modern stochastic gradient methods for optimization, such as Adam \cite{kingma2014adam} and RMSProp \cite{hinton2012neural}, remain unchanged when we multiply the estimated gradients by a fixed quantity; thus, we may harmlessly compute gradients up to multiplicative constants.  

\noindent\textbf{Reverse KL divergence.} We verified in \Cref{eq:ww} that 
\begin{equation*}
    \nabla_{\theta} \mathcal{D}_{KL} [ P_{F} || P_{B} ] = \mathbb{E}_{\tau \sim P_{F}(s_{o}, \cdot)} \left[ \log \frac{p_{F}(\tau | s_{o})}{p_{B}(\tau)} \nabla_{\theta} s_{\theta} (\tau) \right]. 
\end{equation*}
Since $p_{B}(\tau) = p_{B}(\tau | x) \nicefrac{r(x)}{Z}$ and $\mathbb{E}_{\tau \sim P_{F}(s_{o}, \cdot)} \nabla_{\theta} s_{\theta}(\tau) = 0$, the quantity $\nabla_{\theta} \mathcal{D}_{KL}[ P_{F} || P_{B} ]$ may be rewritten as 
\begin{equation*}
    \begin{aligned} 
        \nabla_{\theta} \mathcal{D}_{KL} [ P_{F} || P_{B} ] &= \mathbb{E}_{\tau \sim P_{F}(s_{o}, \cdot)} \left[ \log \frac{p_{F}(\tau | s_{o})}{p_{B}(\tau)} \nabla_{\theta} s_{\theta}(\tau) \right] \\ 
        &+ \mathbb{E}_{\tau \sim P_{F}(s_{o}, \cdot)} \left[ (\log Z) \nabla_{\theta} s_{\theta}(\tau) \right]  \\ 
        &= \mathbb{E}_{\tau \sim P_{F}(s_{o}, \cdot)} \left[ \log \frac{p_{F}(\tau | s_{o})}{p_{B}(\tau | x) r(x)} \nabla_{\theta} s_{\theta}(\tau) \right], 
    \end{aligned} 
\end{equation*}
in which $x$ is the terminal state corresponding to the trajectory $\tau$. Thus proving the statement in \Cref{lemma:aa}. 

\subsection{Proof of \autoref{prop:aaaa}} 

We will derive an expression for the optimal baseline of a vector-valued control variate. For this, let $f$ be the averaged function and $g \colon \tau \mapsto g(\tau)$ be the control variate. Assume, without loss of generality, that $\mathbb{E}_{\pi}[ g ] = 0$ for the averaging distribution $\pi$ over the space of trajectories. In this case, the optimal baseline for the control variate $a^{\star}$ is found by 
\begin{equation}
    a^{\star} = \argmin_{a \in \mathbb{R}} \text{Tr} \left( \text{Cov}_{\tau \sim \pi} [ f(\tau) - a \cdot g(\tau) ]  \right).   
\end{equation}
Thus,  
\begin{equation*}
    \begin{split} 
        a^{\star} = \argmin_{a \in \mathbb{R}} \text{Tr} \bigg(- 2a \cdot \text{Cov}_{\pi} [f(\tau), g(\tau) ] + a^{2} \text{Cov}_{\pi}(g(\tau)) \bigg),   
    \end{split} 
\end{equation*}
which is a convex optimization problem solved by 
\begin{equation}
    \begin{aligned} 
        a^{\star} &= \frac{\text{Tr} \left( \text{Cov}_{\pi} [ f(\tau), g(\tau) ]\right)}{\text{Tr} \left( \text{Cov}_{\pi}[ g(\tau) ] \right)} \\ 
        &= \frac{\text{Tr} \ \mathbb{E}_{\pi} [(f - \mathbb{E}_{\pi} [f(\tau)]) g(\tau)^{T} ]} { \text{Tr} \ \mathbb{E}_{\pi} [ g(\tau) g(\tau)^{T} ]} \\ 
        &= \frac{\mathbb{E}_{\pi} [\text{Tr} \ (f - \mathbb{E}_{\pi} [f(\tau)])^{T} g(\tau) ]} {\mathbb{E}_{\pi} [\text{Tr} \ g(\tau)^{T} g(\tau) ]} \\ 
        &= \frac{\mathbb{E}_{\pi} [(f - \mathbb{E}_{\pi} [f(\tau)])^{T} g(\tau) ]} {\mathbb{E}_{\pi} [g(\tau)^{T} g(\tau) ]}, 
    \end{aligned} 
\end{equation}
in which we used the circular property of the trace. This equation exactly matches the result in \Cref{prop:aaaa}. In practice, we use $g(\tau) = \nabla_{\theta} \log p_{F}(\tau)$ for both the reverse KL- and $\alpha$-divergences, rendering a baseline $a^{\star}$ that depends non-linearly on the sample gradients and is hence difficult to compute in a GPU-powered autodiff framework efficiently. We thus use \Cref{eq:aaa} to estimate $a^{\star}$. 

\section{Additional experiments} 
\label{app:additional}

\begin{figure*}
    \centering
    \includegraphics[width=\textwidth]{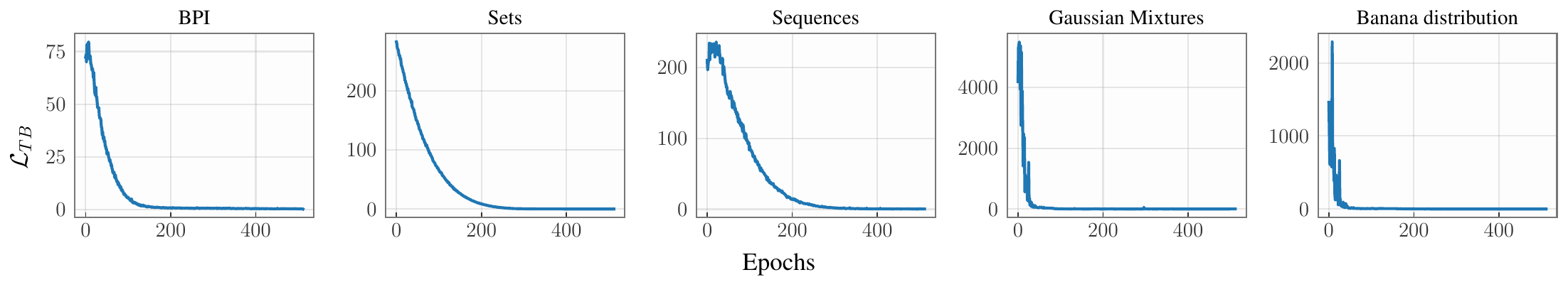}
    \caption{\textbf{Learning curves for a GFlowNet trained by minimizing the TB loss.} The curves' smoothness highlights the low variance of the optimization steps incurred by the stochastic gradients of $\mathcal{L}_{TB}$, which do \emph{not} use a score function estimator.}
    \label{fig:tb}
\end{figure*}

\noindent\textbf{Gradient variance for flow-network-based objectives.} \Cref{fig:tb} shows the learning curve for the TB loss in each of the generative tasks. Notoriously, it underlines the low variance of the optimization steps --- which, contrarily to their divergence-based counterparts, do not rely on a score function estimator --- and suggests that the design of control variates for estimating the gradients of these objectives would not be significantly helpful. Also, the gradient of $\mathcal{L}_{TB}$ depends non-linearly on the score function $\log p_{F}$ and, consequently, it is unclear how to implement computationally efficient variance reduction techniques in this case. 

\begin{wrapfigure}[13]{r}{.3\textwidth} 
    \centering
    \vspace{-18pt} 
    \includegraphics[width=.9\linewidth]{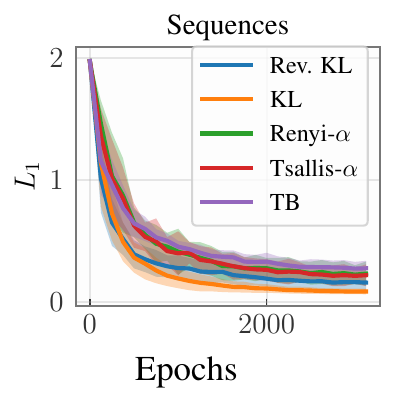}
    \caption{Results for sequence generation with larger batches.} 
    \label{fig:seqs}
\end{wrapfigure}
\noindent\textbf{Forward KL for sequence generation.}
\Cref{fig:conv} shows that alternative approaches in terms of convergence speed outperformed a GFlowNet trained to minimize the forward KL. One possible cause of this underperformance is the high variance induced by the underlying importance sampling estimator. To verify this, we re-run the corresponding experiments, increasing the size of the batch of trajectories for the forward KL estimator to $1024$. \Cref{fig:seqs} presents the experiment's results, with an increased batch size corresponding to an estimator of smaller variance that accelerates the GFlowNet's training convergence. More broadly, this suggests that the design of GFlowNet-specific variance reduction techniques, which we leave to future endeavors, may lead to marked improvements in this family of models.

\section{Experimental details} 
\label{app:details}

The following paragraph provides further implementation details. Regarding open access to the code, we will make the code public upon acceptance. 

\noindent\textbf{Shared configurations.} For every generative task, we used the Adam optimizer \cite{kingma2014adam} to carry out the stochastic optimization, employing a learning rate of $10^{-1}$ for $\log Z_{\theta}$ when minimizing $\mathcal{L}_{TB}$ and $10^{-3}$ for the remaining parameters, following previous works \cite{malkin2022trajectory, pan2023generative, malkin2023gflownets, theory}. We polynomially annealed the learning rate towards $0$ along training, similarly to \cite{sun2023retentive}. Also, we use LeakyReLU \cite{xu2015empirical} as the non-linear activation function of all implemented neural networks.   

\noindent\textbf{Set generation.} We implement an MLP of 2 64-dimensional layers to parameterize the policy's logits $\log p_{F}(s, \cdot)$. We train the model for $512$ epochs with a batch of $128$ trajectories for estimating the gradients. Also, we let $D = 32$ and $N = 16$ be the source's and set's sizes, respectively. 

\noindent\textbf{Autoregressive sequence generation.} We parameterize the logits of the forward policy with a MLP of 2 64-dimensional layers; we pad the sequences to account for their variable sizes. We respectively consider $D = 8$ and $N = 6$ for the source's and sequence's sizes. To approximate the gradients, we rely on a batch of $128$ sequences. 

\noindent\textbf{Bayesian phylogenetic inference.} We parameterize the logits of the forward policy with a 2-layer GIN \cite{xu2018powerful} with a 64-dimensional latent embedding, which is linearly projected to $\log p_{F}$. Moreover, we simulated the JC69 model \cite{Jukes1969} to obtain $25$-sized sequences of nucleotides for each of the $7$ observed species, setting $\lambda = 0.3$ for the instantaneous mutation rate; see \cite{Yang2014} for an introduction to computational phylogenetics and molecular evolution. To estimate the gradients, we relied on batches of 64 trajectories.  

\noindent\textbf{Mixture of Gaussian distributions.} We consider a mixture of $9$ 2-dimensional Gaussian distributions centered at $\mu_{ij} = (i, j)$ for $0 \le i, j \le 2$, each of which having an isotropic variance of $10^{-1}$; see \Cref{fig:alpha}. We use an MLP of 2 64-dimensional layers to parameterize the forward policy. 

\noindent\textbf{Banana-shaped distribution.} The model is specified by \Cref{eq:banana}. We also consider an MLP of 2 64-dimensional layers to parameterize the forward policy. 

\clearpage


\end{document}